%% file: arxiv.tex
\documentclass{article} %
\usepackage{iclr2025_conference,times}

\input{math_commands.tex}

\definecolor{lacamgold5}{RGB}{255, 87, 0}
\newcommand{\rebuttal}[1]{#1}

\usepackage[utf8]{inputenc} %
\usepackage[T1]{fontenc}    %
\usepackage{url}            %
\usepackage{booktabs}       %
\usepackage{amsfonts}       %
\usepackage{nicefrac}       %
\usepackage{microtype}      %
\usepackage{titletoc}       %

\usepackage{enumitem}
\setlist{nolistsep,leftmargin=6ex}
\newcommand{\ourM}{\textsc{DCWM}\xspace}
\newcommand{\our}{\textsc{DC-MPC}\xspace}

\usepackage{xspace}
\newcommand{\eg}{\textit{e.g.\@}\xspace}
\newcommand{\ie}{\textit{i.e.\@}\xspace}

\newcommand{\etc}{\textit{etc.\@}\xspace}

\usepackage{graphicx}
\usepackage{xcolor} %
\usepackage{todonotes}
\usepackage{subcaption}

\usepackage{amsmath}
\usepackage{amssymb}
\usepackage{mathtools}
\usepackage{amsthm}

\usepackage{algcompatible}
\usepackage{algorithm}

\usepackage{listings}

\usepackage{hyperref}       %
\hypersetup{colorlinks=true,allcolors=blue}

\renewcommand{\mid}{\,|\,}

\usepackage[capitalize,nameinlink]{cleveref}
\crefname{section}{Sec.}{Secs.}
\crefname{algorithm}{Alg.}{Algs.}
\crefname{appendix}{App.}{Apps.}
\crefname{definition}{Def.}{Defs.}
\crefname{table}{Table}{Tables}

\makeatletter
\renewcommand*{\@fnsymbol}[1]{\dagger}
\makeatother

\usepackage{wrapfig}

\definecolor{color0}{HTML}{F0F0F5} %
\definecolor{color1}{HTML}{92C5DE} %
\definecolor{color2}{HTML}{B3DE69} %
\definecolor{color3}{HTML}{FC8D59} %
\definecolor{color4}{HTML}{FFD92F} %
\definecolor{color5}{HTML}{D9C4E2} %
\definecolor{color6}{HTML}{E78AC3} %

\usepackage{tikz}
\usetikzlibrary{shapes.geometric,positioning,calc}
\usepackage{tikz-3dplot}

\newcommand{\codebook}[1]{\tikz[baseline=-.5ex, rounded corners=0pt]{%
\foreach \c [count=\i] in {1,...,#1}
    \node[fill=color\c,minimum height=5pt,minimum width=5pt,inner sep=0,draw=black!80,line width=.8pt] at (\i*5pt,0) {};
}}

\newcommand{\histogram}[1]{\tikz[baseline=-.5ex, rounded corners=0pt]{%
\foreach \x [count=\i] in {#1}
    \node[anchor=west,fill=white,minimum height=3pt,minimum width=\x pt,inner sep=0,draw=black!80,line width=.8pt] at (0,\i*3pt) {};
\node[anchor=north east,rotate=-90,minimum width=3pt,font=\tiny,inner sep=0pt,minimum height=1.5em] at (3pt,0) {code idx};}}

\newcommand{\stopgrad}{\tikz[baseline]{
  \draw[thick,black] (-3pt,-1pt) -- (3pt,5pt);
  \draw[thick,black] (-3pt,-4pt) -- (3pt,2pt);}}

\newlength{\nodedist}
\newlength{\nodedistv}

\definecolor{cube}{HTML}{f7f75e}
\newcommand{\cube}{%
\resizebox{1.5em}{1.4em}{
\begin{tikzpicture}[line width=.5pt,baseline=-1ex,cube/.style={very thick,black},
            grid/.style={very thin,gray},
            axis/.style={->,black,thick}]
 \begin{scope}[every node/.append style={yslant=-0.5},yslant=-0.5]
 [cube/.style={very thick,black},
            axis/.style={->,blue,thick}]
   \shade[right color=cube, left color=black!50!cube] (0,0) rectangle +(3,3);
   \node at (0.5,2.5) {};
   \node at (1.5,2.5) {};
   \node at (2.5,2.5) {};
   \node at (0.5,1.5) {};
   \node at (1.5,1.5) {};
   \node at (2.5,1.5) {};
   \node at (0.5,0.5) {};
   \node at (1.5,0.5) {};
   \node at (2.5,0.5) {};
   \draw (0,0) grid (3,3);
 \end{scope}

 \begin{scope}[every node/.append style={yslant=0.5},yslant=0.5]
   \shade[right color=gray!70,left color=cube] (3,-3) rectangle +(3,3);
   \node at (3.5,-0.5) {};
   \node at (4.5,-0.5) {};
   \node at (5.5,-0.5) {};
   \node at (3.5,-1.5) {};
   \node at (4.5,-1.5) {};
   \node at (5.5,-1.5) {};
   \node at (3.5,-2.5) {};
   \node at (4.5,-2.5) {};
   \node at (5.5,-2.5) {};
   \draw (3,-3) grid (6,0);
 \end{scope}

 \begin{scope}[every node/.append style={
     yslant=0.5,xslant=-1},yslant=0.5,xslant=-1
   ]
   \shade[bottom color=cube, top color=black!80] (6,3) rectangle +(-3,-3);
   \node at (3.5,2.5) {};
   \node at (3.5,1.5) {};
   \node at (3.5,0.5) {};
   \node at (4.5,2.5) {};
   \node at (4.5,1.5) {};
   \node at (4.5,0.5) {};
   \node at (5.5,2.5) {};
   \node at (5.5,1.5) {};
   \node at (5.5,0.5) {};
   \draw (3,0) grid (6,3);
 \end{scope}
\end{tikzpicture}}}

\title{Discrete Codebook World Models for \\ Continuous Control}
\author{Aidan Scannell\thanks{Work done while at Aalto University \quad $^*$Equal contribution}\\ %
University of Edinburgh\\
\texttt{aidan.scannell@ed.ac.uk} \\
\And
Mohammadreza Nakhaei$^*$ \\%
Aalto University\\
\And
Kalle Kujanpää$^*$\\
Aalto University\\
\And
Yi Zhao \\
Aalto University\\
\And
Kevin Sebastian Luck \\
Vrije Universiteit Amsterdam \\
\And
Arno Solin \\
Aalto University\\
\And
Joni Pajarinen \\
Aalto University
}

\renewcommand{\paragraph}[1]{{\bfseries #1}~~}

\iclrfinalcopy %
\begin{document}

\maketitle

\begin{abstract}
In reinforcement learning (RL), world models serve as internal simulators, enabling agents to predict environment dynamics and future outcomes in order to make informed decisions.
While previous approaches leveraging discrete latent spaces, such as DreamerV3, have demonstrated strong performance in discrete action settings and visual control tasks, their comparative performance in state-based continuous control remains underexplored. In contrast, methods with continuous latent spaces, such as TD-MPC2, have shown notable success in state-based continuous control benchmarks.
In this paper, we demonstrate that modeling discrete latent states has benefits over continuous latent states
and that discrete codebook encodings are more effective representations for continuous control, compared to alternative encodings, such as one-hot and label-based encodings.
Based on these insights, we introduce DCWM: \textbf{D}iscrete \textbf{C}odebook \textbf{W}orld \textbf{M}odel, a self-supervised world model with a
discrete and stochastic latent space, where latent states are codes from a codebook.
We combine DCWM with decision-time planning to get our model-based RL algorithm, named \our: \textbf{D}iscrete \textbf{C}odebook \textbf{M}odel \textbf{P}redictive \textbf{C}ontrol, which performs competitively against recent state-of-the-art algorithms, including TD-MPC2 and DreamerV3, on continuous control benchmarks.
See our project website \href{www.aidanscannell.com/dcmpc}{\url{www.aidanscannell.com/dcmpc}}.
\looseness-1

\end{abstract}

\section{Introduction}
In model-based reinforcement learning (RL), world models \citep{haRecurrentWorldModels2018}
have been introduced in order to simulate or predict the environment's dynamics in a data-driven way.
An agent equipped with a world model can make predictions about its environment by ``simulating'' possible actions within the model and ''imagining'' the outcomes.
This equips the agent with the ability to plan and anticipate outcomes given a (learned) reward function, and the additional ability to envision transitions and outcomes before taking them in the real world can in turn improve sample efficiency.

One of the state-of-the-art world models, DreamerV2/V3 \citep{hafnerMasteringAtariDiscrete2022,hafner2023mastering} achieves strong performance
in a wide variety of tasks, by ``imagining'' sequences of future states within a world model and using them to improve
their policies.
Interestingly, DreamerV2/V3 introduced a discrete latent space, in the form of a one-hot encoding,
which offered significant benefits over its predecessor, DreamerV1 \citep{hafner2019dream}.
This suggests that discrete latent spaces may have benefits over continuous latent spaces.
It could be from the discrete latent space helping avoid compounding errors over multi-step time horizons or
enabling policy and value learning to harness the benefits of discrete variable processing
for efficiency and interoperability.
In the context of generative modeling, discrete codebooks have been at the heart of many successful approaches
\citep{changMuseTextToImageGeneration2023,esserTamingTransformersHighResolution2021,rameshZeroShotTexttoImageGeneration2021}.
However, in the context of continuous control, TD-MPC2 \citep{hansenTDMPC2ScalableRobust2023}
uses a continuous latent space and significantly outperforms DreamerV3.
Whilst there are multiple differences between TD-MPC2 and DreamerV2/V3, in this paper, we are specifically interested in exploring
if discrete latent spaces can offer benefits for continuous control.
\looseness-3

Recently, \citet{farebrotherStopRegressingTraining2024} showed that training value functions with classification may have benefits over training with regression.
The benefits may arise because {\em (i)} classification considers uncertainty during training (via the cross-entropy loss), {\em (ii)} the categorical distribution is multi-modal so it can consider multiple modes during training, or {\em (iii)} learning in discrete spaces is more efficient.
In the context of world models, it is natural to ask, what benefits are obtained by {\em (i)}
using discrete vs continuous latent spaces
and {\em (ii)} modeling  deterministic vs stochastic transition dynamics.
Further to this, when considering stochastic latent transition dynamics, what is the effect of modeling with {\em (i)} unimodal distributions (\eg Gaussian in continuous latent spaces) vs {\em (ii)} multimodal distributions (\eg categorical in discrete latent spaces).
In this paper, we explore these ideas in the context of world models for model-based RL, \ie does learning a discrete latent space
using classification have benefits over learning a continuous latent space using regression.

\paragraph{Contributions} The main contributions are as follows:
\begin{itemize}
    \item[(C1)] In the context of continuous control, we show that learning discrete latent spaces with classification does have benefits over learning continuous latent spaces with regression.
    \item[(C2)] We show that formulating a discrete latent state using codebook encodings has benefits over alternatives, such as one-hot (like DreamerV2/V3) and label encodings.
    \item[(C3)] Based on our insights, we introduce Discrete Codebook World Model (\ourM): a world model with a discrete latent space
where each latent state is a discrete code from a codebook. It obtains strong performance in the difficult locomotion tasks from DeepMind Control suite \citep{tassa2018deepmind} and manipulation tasks from Meta-World \citep{yu2019meta}.
\end{itemize}

\section{Related Work}

In this section, we recap world models in the context of model-based RL.
We introduce two competing methods for learning latent spaces {\em (i)} those using observation reconstruction and
{\em (ii)} those using latent state temporal consistency objectives.
We then compare methods that learn continuous latent spaces using regression and those
that learn discrete latent spaces using classification.

\paragraph{World models}
Model-based RL is often said to be more sample-efficient than model-free methods.
This is because it learns a model in which it can reason about the world,
instead of simply trying to learn a policy or a value function to maximize the return \citep{haRecurrentWorldModels2018}.
The world model can be used for planning \citep{allen1983planning,basye1992decision}.
A prominent idea has been to optimize the evidence lower bound of observation and reward sequences to learn world models that
operate on the latent space of a learned Variational Autoencoder  (VAE, \citet{kingmaAutoEncoding2014, igl2018deep}).
These models rely on maximizing the conditional observation likelihood $p(\vo_t | \vz_t$), \ie the
reconstruction objective.
The latent space of the model can then be used for both policy learning in the imagination of the world model, known as offline planning, \eg Dreamer
\citep{hafner2019dream}, or for decision-time planning
\citep{rubinstein1997optimization, hafnerLearning2019,schrittwieserMastering2020}.

\paragraph{Latent-state consistency}
Using the reconstruction loss for learning latent state representations is unreliable \citep{lutter2021learning}
and can have a detrimental effect on the performance of model-based methods in various
benchmarks \citep{kostrikov2020image, yarats2021mastering}.
To this end, TD-MPC \citep{hansenTemporalDifferenceLearning2022} and its successor, TD-MPC2 \citep{hansenTDMPC2ScalableRobust2023},
use a consistency loss to learn representations for planning with Model Predictive Path Integral (MPPI) control together with reward and value functions learned through temporal difference methods \citep{williams2015model}.
Note that many prior works learn latent state representations using variants of a self-supervised latent-state consistency objective
\citep{schwarzerDataEfficientReinforcementLearning2020,wangDenoisedMDPsLearning2022,ghugareSimplifyingModelbasedRL2022a,lecunPathAutonomousMachine,georgiev2024pwm,scannell2024iqrl,zhaoSimplifiedTemporalConsistency2023,scannellQuantizedRepresentationsPrevent2024}.
Given the success of learning representations without observation reconstruction in continuous control tasks, we
predominantly focus on this class of methods, \ie methods that use latent-state consistency losses.

\paragraph{Discrete latent spaces}
DreamerV1 \citep{hafner2019dream}, DreamerV2 \citep{hafnerMasteringAtariDiscrete2022}, and DreamerV3 \citep{hafner2023mastering},
are world model methods which learn policies using imagined transitions from their world models. 
They utilize observation reconstruction when learning their world models and perform well across a wide variety of tasks. 
However, they are significantly outperformed by TD-MPC2 in continuous control tasks, which does not reconstruct observations. 
Of particular interest in this paper, is that DreamerV2/V3 introduced a discrete latent space,
in the form of a one-hot encoding, and trained it with a classification objective, significantly improving performance.
In contrast, TD-MPC2 learns a continuous latent space with mean squared error regression.
In this paper, we are interested in learning discrete latent spaces with classification, however, in contrast to DreamerV2/V3,
we seek to avoid observation reconstruction -- due to its poor performance in continuous control (see \cref{fig:decoder-ablation}) -- and instead
learn the latent space using a self-supervised
latent-state consistency loss.\looseness-2

\begin{figure}[t!]
  \small
  \resizebox{\textwidth}{!}{%
  \begin{tikzpicture}[line width=1pt]

  \pgfdeclarelayer{background}
  \pgfsetlayers{background,main}

  \newcommand{\sub}[0]{\scalebox{.8}{\ensuremath t}}
  \newcommand{\subs}[1]{\scalebox{.8}{\ensuremath t{+}#1}}

  \tikzstyle{mynode}=[fill=black!5,draw=black!80,rounded corners=1pt,font=\scriptsize,inner sep=0,align=center,line width=.6pt]
  \tikzstyle{trap}=[mynode,fill=color5,trapezium,text width=4em, minimum height=1.4em,
                    trapezium left angle=-70, trapezium right angle=-70,inner sep=2pt]
  \tikzstyle{blob}=[mynode,circle,minimum width=2.2em,minimum height=2.2em,align=center]
  \tikzstyle{polval}=[mynode,minimum width=6em,minimum height=2em,align=center,text=white,fill=black!60,inner sep=2pt]
  \tikzstyle{arr}=[line width=1pt,black,->]
  \tikzstyle{darr}=[line width=1pt,black,<->,densely dotted]
  \tikzstyle{dlc}=[align=center,font=\scriptsize,text width=5em]
  \tikzstyle{main}=[fill=none,draw=none,rectangle,anchor=south,minimum height=5em,minimum width=4em,fill=color0]

  \draw[fill=color0,draw=color0!60!black,rounded corners=4pt] (-.5\nodedist,-1\nodedistv) rectangle (4.5\nodedist,1.5\nodedistv);

  \node[blob,main] (z0) at (0\nodedist,-1.5) {\codebook{5}\\ Latent code\\$\vc_{\sub}\vphantom{\hat\vc_{\subs{1}}}$};
  \node[blob,main] (z1) at (1\nodedist,-1.5) {\histogram{3,5,8,10,3,6,12}\\$p_{\phi}(\hat\vc_{\subs{1}}\mid{\vc}_{\sub},\va_{\sub})$};
  \node[blob,main] (z2) at (2\nodedist,-1.5) {\codebook{5}\\ Latent code\\$\hat\vc_{\subs{1}}$};
  \node[blob,main] (z3) at (3\nodedist,-1.5) {\histogram{2,1,2,3,5,10,12}\\$p_{\phi}(\hat\vc_{\subs{2}}\mid\hat{\vc}_{\subs{1}},\va_{\subs{1}})$};
  \node[blob,main] (z4) at (4\nodedist,-1.5) {\codebook{5}\\ Latent code\\$\hat\vc_{\subs{2}}$};

  \node[trap] (e0) at (0\nodedist,\nodedistv) {Encoder};
  \node[trap] (e1) at (1\nodedist,\nodedistv) {Encoder};
  \node[trap] (e2) at (3\nodedist,\nodedistv) {Encoder};

  \node[blob,fill=none,draw=none,text width=4em,rectangle] (zh1) at ($(z1)!0.5!(e1)$) {Codebook index};
  \node[blob,fill=none,draw=none,text width=4em,rectangle] (zh2) at ($(z3)!0.5!(e2)$) {Codebook index};

  \coordinate (m0) at ($(z0)!.5!(z1)$);
  \node[blob,fill=color1] (a0) at ($(m0)!(zh1)!(m0)$) {$\va_{\sub}$};
  \coordinate (m1) at ($(z2)!.5!(z3)$);
  \node[blob,fill=color1] (a1) at ($(m1)!(zh2)!(m1)$) {$\va_{\subs{1}}$};
  \node[blob,fill=color6] (r0) at ($(a0) + (0,-3.5)$) {$\vr_{\sub}$};
  \node[blob,fill=color6] (r1) at ($(a1) + (0,-3.5)$) {$\vr_{\subs{1}}$};

  \node[mynode,minimum width=5em,minimum height=4em,outer sep=0,node distance=6em,path picture={\node at (path picture bounding box.center){\includegraphics[height=5.5em]{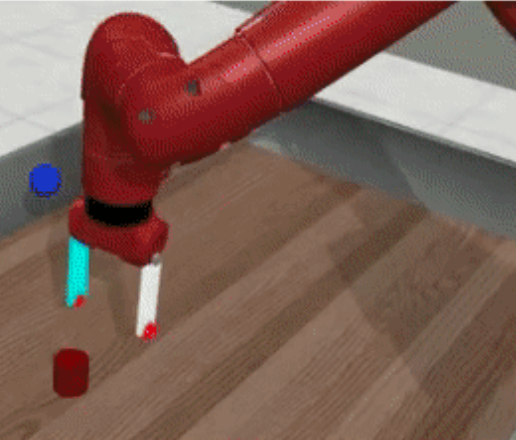}};}] (i0) at (0\nodedist,1.4\nodedistv) {};

  \node[mynode,minimum width=5em,minimum height=4em,outer sep=0,node distance=6em,path picture={\node at (path picture bounding box.center){\includegraphics[height=5.5em]{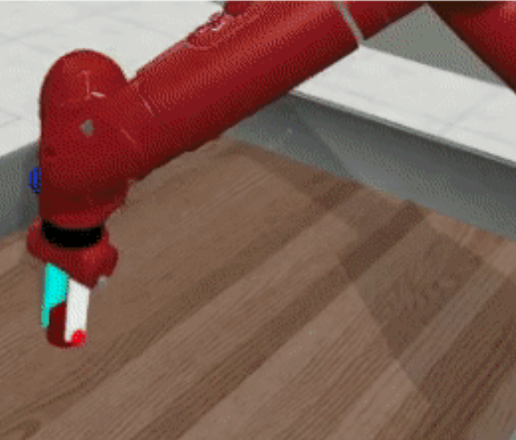}};}] (i1) at (1\nodedist,1.4\nodedistv) {};

  \node[mynode,minimum width=5em,minimum height=4em,outer sep=0,node distance=6em,path picture={\node at (path picture bounding box.center){\includegraphics[height=5.5em]{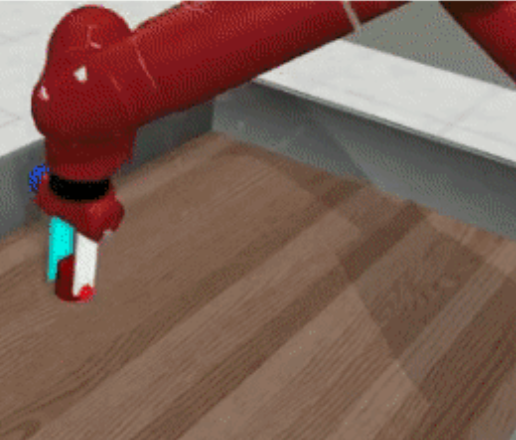}};}] (i2) at (3\nodedist,1.4\nodedistv) {};

  \node[anchor=north,yshift=-1.6em] (o0) at (i0.north) {\color{white}$\vo_{\sub}$};
  \node[anchor=north,yshift=-1.6em] (o1) at (i1.north) {\color{white}$\vo_{\subs{1}}$};
  \node[anchor=north,yshift=-1.6em] (o2) at (i2.north) {\color{white}$\vo_{\subs{2}}$};

  \draw[arr] (a0) to[bend right=20] (r0);
  \draw[arr] (a1) to[bend right=20] (r1);
  \draw[arr] (z0) to[bend right=40] (r0);
  \draw[arr] (z2) to[bend right=40] (r1);
  \node[fill=color0,minimum width=1cm,minimum height=.6cm,opacity=.9] at ($(z0)!.5!(z1) + (0,-4pt)$) {};
  \node[fill=color0,minimum width=1cm,minimum height=.6cm,opacity=.9] at ($(z2)!.5!(z3) + (0,-4pt)$) {};
  \draw[arr] (i0) -- (e0);
  \draw[arr] (i1) -- (e1);
  \draw[arr] (i2) -- (e2);
  \draw[arr] (z0) -- node[below,font=\scriptsize,xshift=0em] {Dynamics} ++(2,0);
  \draw[arr] (z1) -- node[below,text width=5em,font=\scriptsize,xshift=0em,align=center] {ST Gumbel-softmax sampling} ++(2,0);
  \draw[arr] (z2) -- node[below,font=\scriptsize,xshift=0em] {Dynamics} ++(2,0);
  \draw[arr] (z3) -- node[below,text width=5em,font=\scriptsize,xshift=0em,align=center] {ST Gumbel-softmax sampling} ++(2,0);
  \draw[arr] (e0) -- node[right,font=\scriptsize,rotate=90,yshift=6pt] {FSQ} (z0);
  \draw[arr] (e1) -- (zh1);
  \draw[arr] (e2) -- (zh2);
  \draw[arr] (a0) -- (z1);
  \draw[arr] (a1) -- (z3);
  \draw[darr] (zh1) -- node[right,font=\scriptsize,align=left]{Cross-entropy \\ loss} (z1);
  \draw[darr] (zh2) -- node[right,font=\scriptsize,align=left]{Cross-entropy \\ loss} (z3);
  \draw[arr,-] (z4) -- ++(1,0);

  \node[circle,fill=color0,inner sep=0, scale=.7] at ($(e0)!.5!(z0)$) {\cube};
  \node[circle,fill=color0,inner sep=0, scale=.7] (c1) at ($(e1)!.5!(zh1)$) {\cube};
  \node[circle,fill=color0,inner sep=0, scale=.7] (c2) at ($(e2)!.5!(zh2)$) {\cube};
  \node at ($(e1)!.5!(c1)$) {\stopgrad};
  \node at ($(e2)!.5!(c2)$) {\stopgrad};
\end{tikzpicture}}
  \caption{\textbf{World model training} \ourM is a world model with a discrete latent space where each latent state is a discrete code $\vc$ (\protect\codebook{6}) from a codebook $\mathcal{C}$. Observations $\vo$ are first mapped through the encoder and then quantized (\scalebox{.5}{\protect\cube}) into one of the discrete codes. We model probabilistic latent transition dynamics $p_{\phi}(\vc' \mid \vc, \va)$ as a classifier such that it captures a potentially multimodal distribution over the next state $\vc'$ given the previous state $\vc$ and action $\va$. During training, multi-step predictions are made using straight-through (ST) Gumbel-softmax sampling such that gradients backpropagate through time to the encoder. Given this discrete formulation, we train the latent space using a classification objective, \ie cross-entropy loss. Making the latent representation stochastic and discrete with a codebook contributes to the very high sample efficiency of \our.}
  \label{fig:overview}
  \vskip -0.15in
\end{figure}

\rebuttal{
\section{Preliminaries}
\label{sec:preliminaries}

In this section, we recap different types of discrete encodings and compare their pros and cons.
First, let us assume we have three discrete categories: $A,B$, and $C$.
\begin{itemize}
\item \textbf{One-hot encoding} Given categories $A$, $B$, and $C$, a one-hot encoding would take the form $e(A)=[1,0,0]$, $e(B)=[0,1,0]$, and $e(C)=[0,0,1]$ respectively.
\item \textbf{Label encoding} Given categories $A$, $B$, and $C$, label encoding would result in $e(A)=1$, $e(B)=2$, and $e(C)=3$ respectively.
\item \textbf{Codebook encoding} Given categories $A$, $B$, and $C$, a codebook might encode them as $e(A)=[-0.5,-0.5]$, $e(B)=[0,0]$, and $e(C)=[0.5,0.5]$ respectively.
\end{itemize}

\textbf{Ordinal relationships} If we have an ordinal relationship $A < B < C$, label and codebook encodings can ensure $|e(A) - e(B)| < |e(A) - e(C)|$, where $e(\cdot)$ is the encoding function.
In this case, the global ordering is preserved along both dimensions of the codebook.
It is worth noting that codebook encodings are flexible enough to model ordinal relationships in multiple dimensions.
For example, the following code vectors exhibit opposite ordering along their two dimensions $e(E)=[0.5,-0.5]$, $e(F)=[0, 0]$, $e(G)=[-0.5,0.5]$, which adds a level of modeling flexibility.
One-hot encoding, however, results in $|e(A) - e(B)| = |e(A) - e(C)| = \sqrt{2}$ for all distinct pairs, eliminating any notion of ordering.
Whilst this may be beneficial in some scenarios, \eg, when modeling distinct categories like fruits,
it means that they cannot capture the inherent ordering in continuous data.

\textbf{Sparsity and dimensionality} Another downside of one-hot encodings is that they create sparse data (\ie, data with many zero values),
which can have a negative impact on neural network training.
In contrast, label and codebook encodings create dense data (\ie many non-zero values).
Finally, it is worth noting that one-hot encodings have high dimensionality, especially when there are many categories.
This makes them memory-intensive and slow to train when using a large number of categories.\looseness-1

In this work, we show that discrete codebook encodings resulting from quantization
\citep{mentzerFiniteScalarQuantization2023}
offer benefits over both one-hot and label encodings when learning discrete latent spaces for continuous control.
This is because they preserve ordinal relationships in multiple dimensions whilst being simpler,
much lower-dimensional and having less memory requirements.

}

\section{Method}
\label{sec:method}

In this section, we detail our method, named \textit{Discrete Codebook Model Predictive Control} (\our),
which is a model-based RL algorithm which {\em (i)}~learns a world model with a discrete latent space,
named \textit{Discrete Codebook World Model} (\ourM), and then,
{\em (ii)}~performs decision-time planning with MPPI.
The paper's main contribution is formulating a discrete latent space using quantization such that latent states are codes from a codebook.
This allows us to train the latent representation using classification, in a self-supervised manner.
See \cref{fig:overview} for an overview of \ourM, \cref{alg:world_model_alg} for details of world model training and \cref{alg:mppi_alg} for details on the MPPI planning procedure.\looseness-3

We consider Markov Decision Processes (MDPs, \cite{bellmanMarkovianDecisionProcess1957}) $\mathcal{M} = (\mathcal{O}, \mathcal{A}, \mathcal{P}, \mathcal{R}, \gamma)$,
where agents receive observations $\vo_{t} \in \mathcal{O}$ at time step $t$, perform actions $\va_{t} \in \mathcal{A}$, and
then obtain the next observation $\vo_{t+1} \sim \mathcal{P}(\cdot \mid \vo_{t}, \va_{t})$ and reward $r_{t} = \mathcal{R} (\vo_{t}, \va_{t})$.
The discount factor is denoted $\gamma \in [0, 1)$.\looseness-2

\subsection{World model}
\label{sec:world-model-training}
Learning world models with discrete latent spaces (\eg DreamerV2) has proven powerful in a wide variety of domains.
However, these approaches generally perform poorly in continuous control tasks when compared to algorithms like TD-MPC2 and
TCRL \citep{zhaoSimplifiedTemporalConsistency2023}, which use continuous latent spaces.
Rather than representing a discrete latent space using a one-hot encoding, as was done in DreamerV2, \our aims to
construct a more expressive representation which is effective for continuous control.
More specifically, \our represents discrete latent states as codes from a discrete codebook, obtained via
finite scalar quantization (FSQ, \citet{mentzerFiniteScalarQuantization2023}).
The world model can subsequently benefit from the advantages of discrete representations,
\eg efficiency and training with classification, whilst performing well in continuous control tasks.

\paragraph{Components}
\our has six main components:
\begin{align}
&\text{Encoder:} & \vx &= e_{\theta} (\vo) \in \R^{|\mathcal{L}| \times d} \label{eq:encoder} \\
&\text{Latent quantization:} & \vc &= f(\vx) \in \mathcal{C} \label{eq:fsq} \\
&\text{Dynamics:} & \vc' &\sim \mathrm{Categorical} \left( p_{1},\ldots,p_{|\mathcal{C}|} \right) \quad \text{with }
p_{i} = p_{\phi}({\vc}'=\vc^{(i)} \mid \vc, \va) \\ %
&\text{Reward:} & r &= R_{\xi}(\vc, \va) \in \R \\
&\text{Value:} & \vq &= \mathbf{q}_{\psi} (\vc, \va) \in \R^{N_{q}} \label{eq:value} \\
&\text{Policy prior:} & \va &= \pi_{\eta} (\vc) \label{eq:policy}
\end{align}
The encoder $e_{\theta}(\cdot)$ first maps observations $\vo$ to continuous latent vectors $\vx \in \R^{b \times d}$, where the number of channels $b$ and the latent dimension $d$ are hyperparameters.
This continuous latent vector $\vx$ is then quantized $f(\cdot)$ into one of the discrete latent codes $\vc \in \mathcal{C}$
from the (fixed) codebook $\mathcal{C}$, using
finite scalar quantization (FSQ, \citet{mentzerFiniteScalarQuantization2023}).
\rebuttal{
As we have a discrete latent space, we formulate the transition dynamics to model the distribution over
the next latent state $\vc'$ given the previous latent state $\vc$ and action $\va$.
That is, we model stochastic transition dynamics in the latent space.
We denote the probability of the next latent state $\vc'$ taking the
value of the $i^{\text{th}}$ code $\vc^{(i)}$ as $p_{i} = p_{\phi}(\vc'=\vc^{(i)} \mid \vc, \va)$.
This results in the next latent state following a categorical distribution $\vc' \sim \mathrm{Categorical} \left( p_{1},\ldots,p_{|\mathcal{C}|} \right)$.
We use a standard classification setup, where
we use an MLP to predict the logits $\vl = \{l_{1}, \ldots, l_{|\mathcal{C}|}\} = d_{\phi} (\vc, \va) \in \R^{|\mathcal{C}|}$.
Note that logits are the raw outputs from the final layer of the neural network (NN), which represent the unnormalized probabilities
of the next latent state $\vc'$ taking the value of each discrete code in the codebook $\mathcal{C}$.
The logit for the $i^{th}$ code is given by $l_{i} = [d_{\phi} (\vc, \va)]_{i} \in \R$.
We then apply softmax to obtain the probabilities $\{p_{i} \}_{i=1}^{|\mathcal{C}|}$ of the next latent state taking each
discrete code in the codebook $\mathcal{C}$, \ie, $p_{i} = \softmax_{i}(\vl)$.
\our utilizes the discrete codes $\vc$ as its latent state for future predictions and decision-making.
}

\paragraph{Quantized latent space}
\input{figs/hypercube_figure}
\ourM uses a discretized latent space where world states are encoded as discrete codes
from a codebook $\mathcal{C}$.
We use latent quantization to
enforce data compression and encourage organization \citep{hsuDisentanglementLatentQuantization2023}.
However, we implement this using finite scalar quantization (FSQ, \citet{mentzerFiniteScalarQuantization2023})
instead of dictionary learning \citep{vandenoordNeuralDiscreteRepresentation2017}.
As a result, our codebook is fixed and we obviate two codebook learning loss terms, which stabilizes early training.
In this section, we will give an overview of our discretization method which utilizes codebooks.
First, let us assume the output of the encoder is a tensor\footnote{For simplicity, we omit here the batch dimension.} $\mathbf{x} \in \mathbb{R}^{b \times d}$, with $d$ dimensions and $b$ as the number of channels.\looseness-1

Each latent dimension is quantized into a codebook $\mathcal{C}$.
That is, we have $d$ independent codebooks, one for each latent dimension.
Our first step is to define the size of the codebook for each dimension, \ie to define the ordered set of quantization levels $\mathcal{L} = \left\lbrace L_1, L_2, \cdots, L_b \right\rbrace$.
Each quantization level $L_i$ corresponds to the $i\text{-th}$ channel, \eg $L_1$ defines the number of discrete values in the first channel, $L_2$ for the second and so on.
In short, a quantization level of \eg $L_i=11$ would mean that we discretize each dimension in the $i\text{-th}$ channel into 11 distinct values/symbols.
We use integers as symbols, which would mean that the code for \rebuttal{dimension $d$ in channel $i$} would be a symbol from the set $\lbrace -5, -4, \cdots, 0, \cdots, 4, 5\rbrace$.
In practice, for fast conversion from continuous values to codes we use a similar discretization scheme as FSQ and apply the function\looseness-1
\begin{align}
f : \vx, \mathcal{L}, i \rightarrow \mathrm{round}\left(\left\lfloor \frac{L_i}{2}\right\rfloor \cdot  \text{tanh}(\vx_{i,:})\right),
\label{Eq::discretization}
\end{align}
to each channel, taking the output $\vx$ of the encoder and the channel quantization level $L_i$.
This approach results in a codebook with $|\mathcal{C}| = \prod_{i=1}^{b} L_{i}$ unique codes for each dimension $d$, each code being made of $b$ symbols, \ie a $b\text{-dimensional}$ vector.

Intuitively, this results in a Voronoi partition of the $b$-dimensional space
in each dimension $d$, where any point in space is assigned to one of the equidistantly placed centroids via \cref{Eq::discretization}.
See \cref{fig:fsq} for a visualization.
In effect, this leads to an efficient and fast discretization of the latent embedding space.\looseness-3

In practice, \cref{Eq::discretization} is not differentiable.
To solve this for using standard deep learning libraries, we use the straight-through gradient estimation (STE) approach with $\mathrm{round\_ste}(\vx) : x \rightarrow x + \mathrm{sg}(\mathrm{round}(\vx)-\vx)$, where the function $\mathrm{sg}(\cdot )$ stops the gradient flow.
Furthermore, we normalize codes to be in the range $[-1,1]$ after the discretization step
\rebuttal{as improved performance was reported by \citet{mentzerFiniteScalarQuantization2023}.}
The hyperparameters of this approach are the number of channels $b$ and the number of code symbols per channel $L_i$, \ie quantization levels.
In our experiments, we found the quantization levels $\mathcal{L}=\lbrace5,3\rbrace$ (\ie $b=2$ channels) to be sufficient.

\paragraph{World model training}
We train our world model components $e_{\theta}, d_{\phi}, R_{\xi}$ jointly using backpropagation through time (BPTT) with the following objective
\begin{align} \label{eq:world-model-loss}
  \mathcal{L}(\theta, \phi, \xi; \mathcal{D})
&= \E_{
  (\vo, \va, \vo',r)_{0:H} \sim \mathcal{D}
  }
  \Bigg[
  \sum_{h=0}^{H}  \gamma^{h} \Big(
  \mathrm{CE}(\underbrace{p_{\phi}(\hat{\vc}_{h+1} \mid \hat{\vc}_{h}, \va_{h}), \vc_{h+1}}_{\text{Latent-state consistency}})
    + \underbrace{\| R_{\xi}(\hat{\vc}_{h}, \va_{h}) - r_{h} \|_{2}^{2}}_{\text{Reward prediction}} \Big) \nonumber
  \Bigg] \\
&\text{with }
\underbrace{\hat{\vc}_{0} = f(e_{\theta}(\vo_{0}))}_{\text{First latent state}} \quad
\underbrace{\hat{\vc}_{h+1} \sim p_{\phi}(\hat{\vc}_{h+1} \mid \hat{\vc}_{h}, \va_{h})}_{\text{Stochastic dynamics}} \quad
\underbrace{\vc_{h} = \mathrm{sg}(f(e_{\theta}(\vo_{h})))}_{\text{Target latent code}},
\end{align}
where $H$ denotes the multi-step prediction horizon and $\gamma$ is the discount factor.
The first predicted latent code $\hat{\vc}_{0}$ is obtained by passing the observation $\vo_{0}$ through the
encoder and then quantizing the output.
At subsequent time steps, the dynamics model predicts the probability mass function over the next latent code
$p_{\phi}(\hat{\vc}_{h+1} \mid \hat{\vc}_{h}, \va_{h})$.
Given this probabilistic dynamics model, we must consider how to make $H\text{-step}$ predictions in the latent space.
In practice, we propagate uncertainty by sampling and we use the straight-through (ST) Gumbel-softmax trick \citep{jang2017categorical,maddison2017the}  so that gradients backpropagate
through our samples to the encoder.
Note that gradients must flow back to the encoder at the first time step when it was used to obtain
the first latent code $\hat{\vc}_{0}$, as the target codes $\vc$ are obtained by passing the
next observation $\vo'$ through the encoder and using the stop gradient operator $\mathrm{sg}$.
We then train our dynamics ``classifier'' using the cross-entropy (CE) loss.
Finally, we note that our reward model $R_{\xi}$ is trained jointly with the encoder $e_{\theta}$ and dynamics model $p_{\phi}$
to ensure that the world model can accurately predict rewards in the latent space.\looseness-2

\paragraph{Policy and value learning}
We learn the policy $\pi_{\eta}(\vc)$ and action-value functions $\vq_{\psi}(\vc,\va)$
in the latent space using the actor-critic RL method TD3 \citep{fujimotoAddressingFunctionApproximation2018}.
However, we follow \citet{yaratsMasteringVisualContinuous2021,zhaoSimplifiedTemporalConsistency2023}
and augment the loss with $N\text{-step}$ returns.
The main difference to TD3 is that instead of using the original observations $\vo$, we map them through the
encoder $\vc = f(e_{{\theta}}(\vo))$ and learn the actor/critic in the discrete latent space $\vc$.
We also reduce bias in the TD target by following REDQ \citep{chenRandomizedEnsembledDouble2021}
and learning an ensemble of $N_{q}=5$ critics\rebuttal{, as was done in TD-MPC2}.
When calculating the TD target we randomly subsample two of the critics and use the minimum of these two.
Let us denote the indices of the two randomly subsampled critics as $\mathcal{M}$.
The critic is then updated by minimizing the following objective:
\begin{align} \label{eq:value-loss}
  \mathcal{L}_{q}(\psi; \mathcal{D}) &= \E_{(\vo, \va, \vo', r)_{n=1}^{N} \sim \mathcal{D}} \left[ \frac{1}{N_{q}} \sum_{k=1}^{N_{q}} (q_{\psi_{k}}(\underbrace{f(e_{{\theta}}(\vo_{t}))}_{\vc_{t}}, \va_{t}) - y)^{2}  \right], \\
  y &= \sum_{n=0}^{N-1} \gamma^{n} r_{t+n} + \gamma^{N} \min_{k \in \mathcal{M}} q_{\bar{\psi}_{k}}(\underbrace{f(e_{{\theta}}(\vo_{t+N}))}_{\vc_{t+N}}, \va_{t+N}),
\quad \text{with} \ \va_{t+n} = \pi_{\bar{\eta}}(\vc_{t+n}) + \epsilon_{t+n}, \nonumber
\end{align}
where we use policy smoothing by adding clipped Gaussian noise $\epsilon_{t+n} \sim \mathrm{clip} \left(\mathcal{N} (0,\sigma^{2}), -c, c \right)$ to the
action $\va_{t+n} = \pi_{\bar{\eta}}(\vc_{t+n}) + \epsilon_{t+n}$.
We then use the target action-value functions $\vq_{\bar{\psi}}$ and the target policy $\pi_{\bar{\eta}}$ to
calculate the TD target $y$.
Note that the target networks use an exponential moving average, \ie $[\bar{\psi}, \bar{\eta}] \leftarrow (1-\tau)[\bar{\psi},\bar{\eta}] + [\psi,\eta]$.
We follow REDQ and learn the actor by minimizing
\begin{align} \label{eq:policy-loss}
 \mathcal{L}_{\pi}(\eta ; \mathcal{D}) = - \E_{\vo_{t} \sim \mathcal{D}} \bigg[ \frac{1}{|\mathcal{M}|} \sum_{\psi_{k \in \mathcal{M}}} q_{\psi_{k}}(\underbrace{f(e_{{\theta}}(\vo_{t}))}_{\vc_{t}}, \pi_{\eta}(\underbrace{f(e_{{\theta}}(\vo_{t}))}_{\vc_{t}})) \bigg].
\end{align}
That is, we train the actor to maximize the average action value over two subsampled critics.

\paragraph{Summary}
Whilst this world model shares some similarities with TD-MPC2, there are some important distinctions.
First, the latent space is represented as a discrete codebook which enables \our to train the dynamics model
using the cross-entropy loss.
Importantly, the cross-entropy loss considers a (potentially multimodal) distribution over the
predicted latent codes during both training and inference.
In contrast, TD-MPC2 considers deterministic dynamics and uses mean squared error regression.
Interestingly, our experiments suggest that our stochastic dynamics model offers benefits in deterministic environments.
Second, \our does not use value prediction when training the encoder.
Instead, we follow the insight from \citet{zhaoSimplifiedTemporalConsistency2023}
that value prediction is not necessary for
obtaining a good latent representation and instead, train the action-value function separately.\looseness-1

Importantly, our discrete latent space is parameterized as a set of discrete codes from a codebook.
\rebuttal{
It is worth highlighting that our codebook encoding preserves ordinal relationships between observations.
This contrasts with one-hot encodings which were used by DreamerV2 \citep{hafnerMasteringAtariDiscrete2022}.
See \cref{sec:preliminaries} for a comparison of the different discrete encodings.\looseness-2
}
We hypothesize that this will offer significant improvements when representing continuous state vectors in a
discrete space.

\subsection{Decision-time planning}
\label{sec:mppi}
\our follows TD-MPC2 and leverages the world model for decision-time planning.
It uses MPC to obtain a closed-loop controller and
uses (modified) MPPI \citep{williams2015model} as the underlying trajectory optimization
algorithm \citep{bettsSurvey1998,scannellTrajectory2021}.
MPPI is a sampling-based trajectory optimization method which does not require gradients.
See \cref{alg:mppi_alg} for full details.
At each environment step, we estimate the parameters $\bm\mu^{*}_{0:H}, \bm\sigma^{*}_{0:H}$ of a diagonal multivariate Gaussian
over a $H\text{-step}$ action sequence that maximizes the following objective
\begin{subequations}\label{eq:mppi}
\begin{align}
\bm\mu^{*}_{0:H}, \bm\sigma^{*}_{0:H} &= \argmax_{\bm\mu_{0:H},\bm\sigma_{0:H}}
  \E_{\va_{0:H} \sim \mathcal{N}\left(\bm\mu_{0:H}, \text{diag}(\bm\sigma^{2}_{0:H})\right)}
\left[J(\va_{0:H}, \vo) \right] \\
 J(\va_{0:H}, \vo) &= \sum_{h=0}^{H-1} \gamma^{h} R_{\xi}(\hat{\vc}_{h}, \va_{h}) + \gamma^{H} \frac{1}{|\mathcal{M}|} \sum_{\psi_{k} \in \mathcal{M}}q_{\psi_{k}}(\hat{\vc}_{H},\va_{H})  \\
  &\text{s.t. } \quad \hat{\vc}_{0}=f(e_{\theta}(\vo)) \quad \text{and} \quad
  \hat{\vc}_{h+1}= \sum_{i=1}^{|\mathcal{C}|} \Pr(\hat{\vc}_{h+1}=\vc^{(i)} \mid \hat{\vc}_{h}, \va_{h}) \vc^{(i)},
\end{align}
\end{subequations}
where $H$ is the planning horizon and $\gamma$ is a discount factor.
MPPI solves \cref{eq:mppi} in an iterative manner. It starts by sampling candidate action sequences and evaluating
them using the objective $ J(\va_{0:H}, \vo)$.
It then refits the sampling distribution's parameters $\bm\mu_{0:H}$, $\bm\sigma^{2}_{0:H}$ based on a weighted average.
After several iterations, we select an action trajectory and apply its first action $\va_{0}^{(i^{*})}$ in the environment.
Note that during training we promote exploration by adding Gaussian noise.
Importantly, \cref{eq:mppi} uses the action-value function $\vq_{\psi}(\vc,\va)$ to bootstrap the planning horizon such
that it estimates the full RL objective.
\our follows TD-MPC2 and \rebuttal{warm starts} the planning process with $N_{\pi}$ action sequences originating from the
prior policy $\pi_{\eta}$ and we warm start by initializing $\bm\mu_{0:H}$, $\bm\sigma^{2}_{0:H}$ as the solution to the previous
time step shifted by one. See  \cref{app:method,alg:mppi_alg} for further details.

Note that at planning time, we do not sample from the transition dynamics $p(\vc_{h+1}\mid \vc_{h}, \va_{h})$
because this introduces unwanted stochasticity.
Instead, we take the expected code, which is a weighted sum over the codes in the codebook.
Whilst the expected value of a discrete variable does not necessarily take a valid discrete value,
we find it effective in our setting.
This is likely because our discrete codes have an ordering such that expected values simply
interpolate between the codes in the codebook.

\section{Experiments}

In this section, we experimentally evaluate \our in a variety of continuous control tasks from the DeepMind Control Suite
(DMControl) \citep{tassa2018deepmind}, Meta-World \citep{yu2019meta} and MyoSuite \citep{MyoSuite2022} against a number of baselines and ablations.
Our experiments seek to answer the following research questions:
\begin{enumerate}
    \item[RQ1] Does \our's discrete latent space offer benefits over a continuous latent space?
    \item[RQ2] What is important for learning a latent space: {\em (i)} classification loss, {\em (ii)} discrete codebook, {\em (iii)} stochastic dynamics or {\em (iv)} multimodal dynamics?
    \item[RQ3] Does \our's codebook offer benefits for dynamics/value/policy learning over alternative discrete encodings such as {\em (i)} one-hot encoding (similar to DreamerV2) and {\em (ii)} label encoding?
    \item[RQ4] How does \our compare to state-of-the-art model-based RL algorithms leveraging latent state embeddings, especially in the hard DMControl and Meta-World tasks?
\end{enumerate}

\paragraph{Experimental Setup}
We compared \our against two state-of-the-art model-based RL baselines, namely DreamerV3 \citep{hafner2023mastering} which utilizes a discrete one-hot encoding as its latent state and TD-MPC2 \citep{hansenTDMPC2ScalableRobust2023} using a continuous latent space.
\rebuttal{We also compare against soft actor-critic (SAC) \citep{haarnojaSoft2018}, a model-free RL baseline, and
  the original TD-MPC \citep{hansenTemporalDifferenceLearning2022}.}
Our proposed approach utilized a latent space with $d=512$ dimensions and $b=2$ channels, with $15$ code symbols per dimension by using FSQ levels $\mathcal{L}= \lbrace L_1=5, L_2=3\rbrace$.\looseness-1

\subsection{Comparison of different latent spaces}
\label{sec:latent-space-ablation}
We first evaluate how different latent dynamics formulations affect the performance.
We seek to answer the following:
{\em (i)} do discrete latent spaces offer benefits over continuous latent spaces?
{\em (ii)} does training with classification (cross-entropy) offer benefits over mean squared error regression? and
{\em (iii)} does modeling stochastic (and potentially multimodal) transition dynamics offer benefits?

\begin{figure}[t]
  \centering
  \begin{subfigure}{0.49\textwidth}
    \centering
    \includegraphics[width=\linewidth]{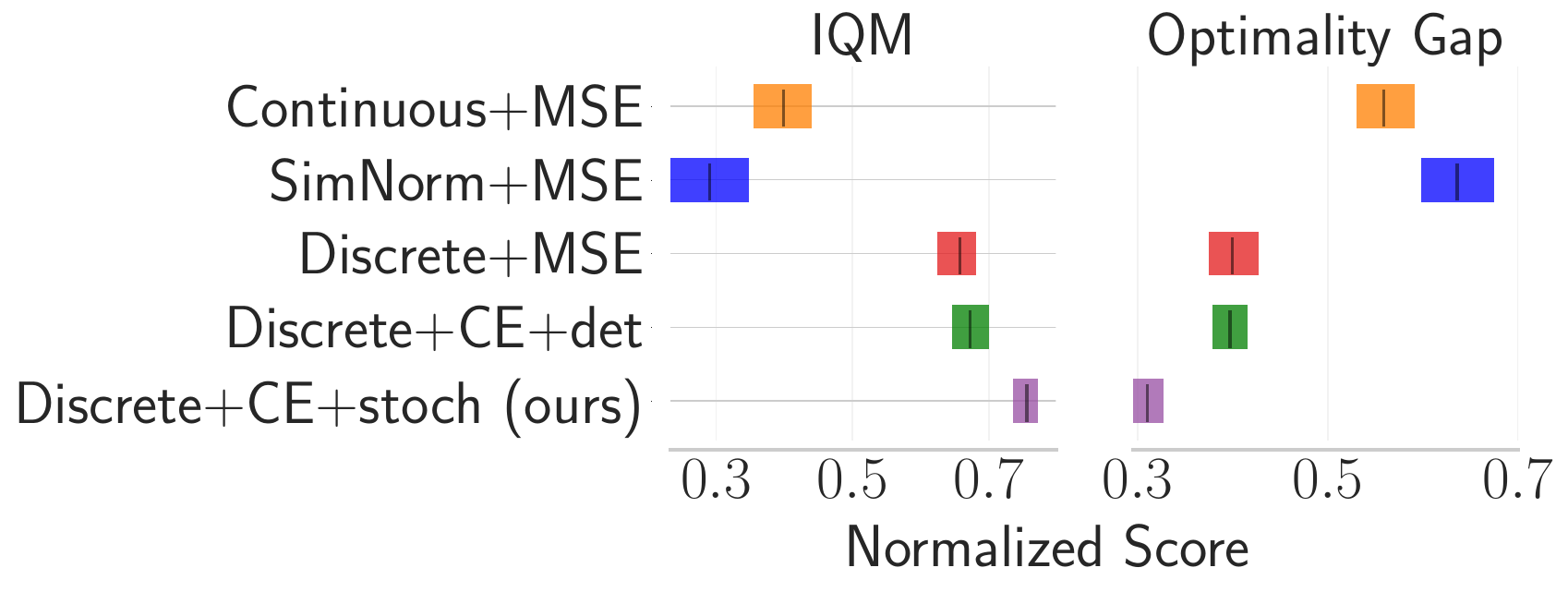}
    \caption{\textbf{Aggregate statistics at 500k environment steps}}
  \end{subfigure}%
  \hfill
  \begin{subfigure}{0.45\textwidth}
    \centering
    \includegraphics[width=\linewidth]{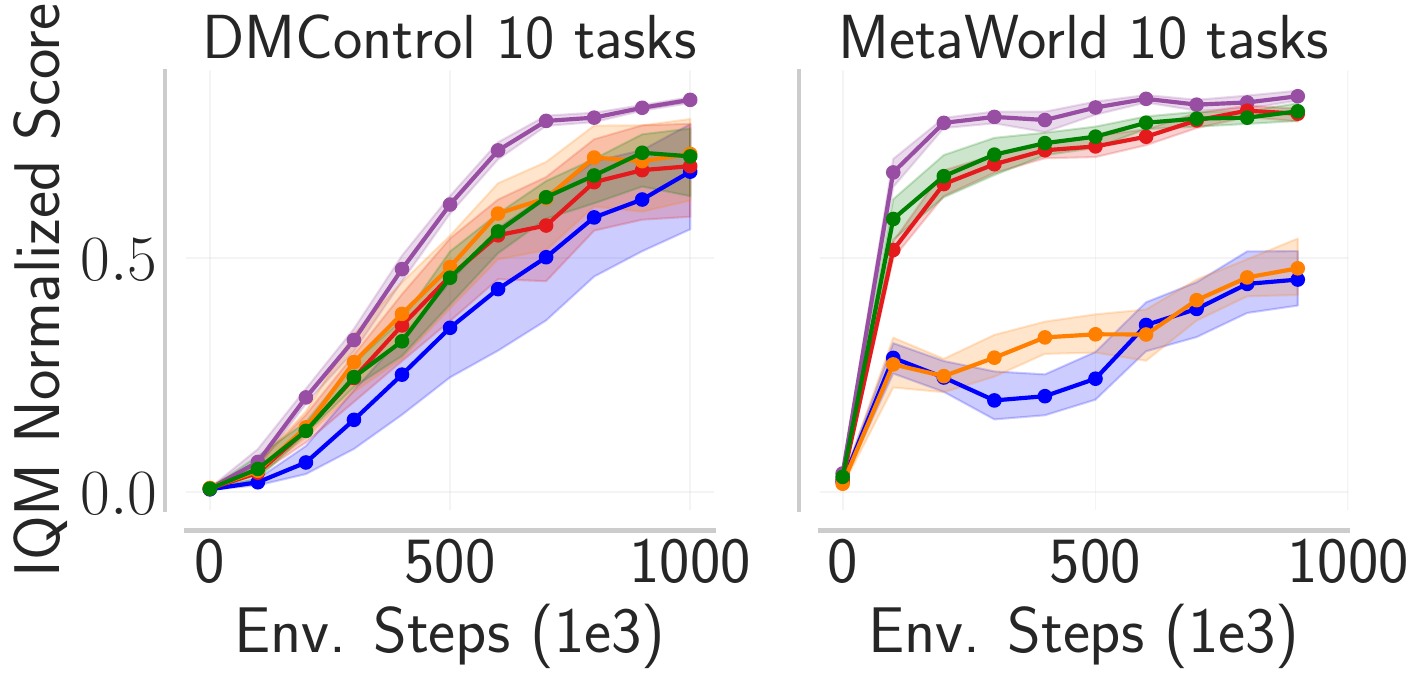}
    \caption{\textbf{Training curves}}
  \end{subfigure}
  \caption{\rebuttal{\textbf{Latent space ablation} Evaluation of {\em (i)} discrete (Discrete) vs continuous (Continuous) latent spaces, {\em (ii)} using cross-entropy (CE) vs mean squared error (MSE) for the latent-state consistency loss, and {\em (iii)} formulating a deterministic (det) vs stochastic (stoch) dynamics model.
      Discretizing the latent space (red) improves sample efficiency over the continuous latent space (orange) and formulating stochastic dynamics and training with cross-entropy (purple) improves performance further.\looseness-1}}
  \label{fig:latent-space-comparison}
  \vskip -0.2in
\end{figure}

\rebuttal{In our experiments, we consider both continuous and discrete latent spaces to investigate the impact of discretizing the latent space of the world model.
In \cref{fig:latent-space-comparison,fig:gaussian-ablation}, the experiments with discrete latent spaces are labelled with ``Discrete'' (red, green, and purple) whilst
continuous latent spaces are labelled ``Continuous'' (orange).
We also evaluate \our using the simplical normalization used in TD-MPC2 -- which bounds the latent space -- labelled ``SimNorm'' (blue) .
Experiments labelled with ``MSE'' were trained with mean squared error regression whilst those labelled ``CE'' were trained with
the cross-entropy classification loss.
The experiment labelled ``Discrete+CE+det'' used FSQ to get a discrete latent space and trained with the cross-entropy loss, where the logits were obtained as
the MSE between the dynamics prediction and each code in the codebook.
This experiment enabled us to test if \our's performance boost resulted from training with the cross-entropy loss or from making the dynamics stochastic.
In \cref{fig:gaussian-ablation}, experiments labelled with ``log-lik.'' were trained by maximizing the log-likelihood,
\ie cross-entropy for ``FSQ-log-lik.'' (purple), Gaussian log prob. for ``Gaussian+log-lik.'' (blue), and Gaussian mixture log prob. for ``GMM+log-lik.'' (green).
}

\paragraph{Discrete vs continuous latent spaces}
The experiments using discrete latent spaces (red and purple) significantly outperform the ones with
continuous latent spaces in terms of sample efficiency.
This suggests that our discrete codebook encoding offers significant benefits over continuous latent spaces.

\paragraph{Classification vs regression}
Interestingly, training a deterministic discrete latent space using MSE regression (red) does not perform as
well as training a stochastic discrete latent space using classification (purple).
However, our experiment with the deterministic discrete latent space using classification (green) confirms that the benefit arises from the stochasticity of
our latent space.
This suggests that using straight-through Gumbel-softmax sampling \citep{jang2017categorical}  when making multi-step dynamics predictions during training boosts performance.
\rebuttal{Our results extending TD-MPC2 to use \our's discrete stochastic latent space in \cref{fig:tdmpc2_with_dcwm_rliable} support this conclusion.}

\paragraph{Deterministic vs stochastic}
Given that modeling a stochastic latent space and training with maximum log-likelihood is beneficial for discrete latent
spaces, we now test if this holds in continuous latent spaces.
To this end, we formulate two stochastic continuous latent spaces and compare them in \cref{fig:gaussian-ablation}.
The first models a unimodal Gaussian distribution (blue) whilst the second models a multimodal
Gaussian mixture model (GMM) (green).
Interestingly, these stochastic transition models sometimes increase sample efficiency on DMControl tasks
when compared to their deterministic counterparts (orange).
However, they drastically underperform on Meta-World tasks.

\rebuttal{Our method (purple) has a discrete latent space, is trained by maximum log-likelihood (\ie cross-entropy), and
  models a (potentially multimodal) distribution over the latent transition dynamics during training.}
These factors, combined with using ST Gumbel-softmax sampling, offer improved
sample efficiency over continuous latent spaces.

\begin{figure}[t]
  \centering
\includegraphics[width=0.95\textwidth]{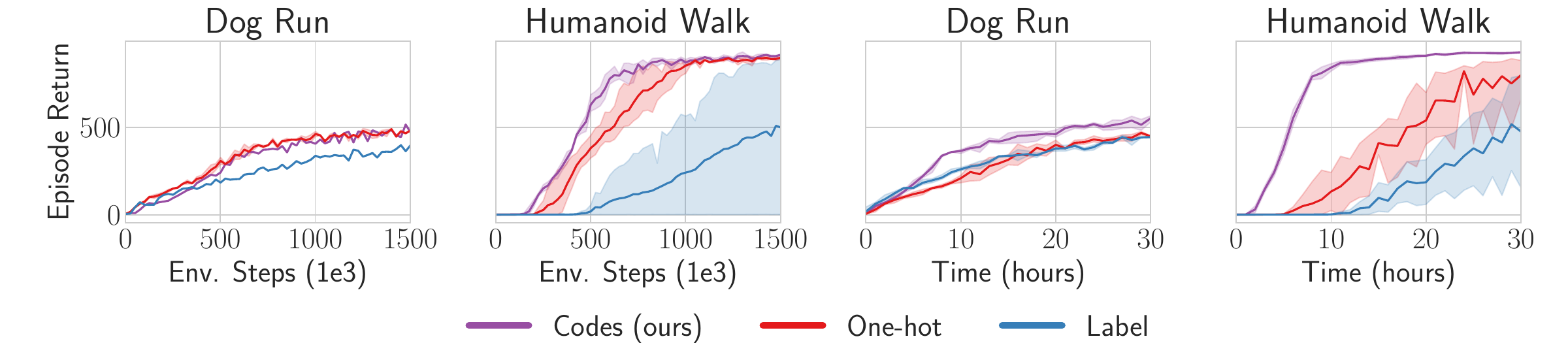}
\caption{\rebuttal{\textbf{Discrete encodings ablation} \our with its discrete codebook encoding (purple) outperforms using \our with one-hot encoding (red) and label encoding (blue), in terms of both sample efficiency (left) and computational efficiency (right).
Dynamics model used codes $p_{\phi}(\mathbf{c}' \mid \mathbf{c}, \mathbf{a})$ whilst reward $R_{\xi}(\mathbf{e}, \mathbf{a})$, critic $Q_{\psi}(\mathbf{e}, \mathbf{a})$ and prior policy $\pi_{\eta}(\mathbf{e})$ used the respective encoding $\mathbf{e}$.}
}
\label{fig:discrete-encodings-comparison}
\vspace{-1.2em}
\end{figure}

\subsection{Impact of latent space encoding}
\rebuttal{
  Our world model consists of NNs for the dynamics $p_{\phi}(\mathbf{c}' \mid \mathbf{c}, \mathbf{a})$, reward $R_{\xi}(\mathbf{c}, \mathbf{a})$, critic $Q_{\psi}(\mathbf{c}, \mathbf{a})$, and prior policy $\pi_{\eta}(\mathbf{c})$,
  which all make predictions given the discrete codebook encoding $\mathbf{c}=\mathbf{e}_{\text{codes}}$.
In \cref{fig:discrete-encodings-comparison}, we evaluate what happens when we replace the codebook encoding $\mathbf{c}$
with {\em (i)} label encoding $e_{\text{label}} = i \in \{1,\dots,|\mathcal{C}|\}$
and {\em (ii)} one-hot encoding $\ve_{\text{one-hot}} =\vv \in \{0,1\}^{|\mathcal{C}|}$ given $\textstyle \sum_{i=1}^{|\mathcal{C}|} v_{i} =1$.
In these experiments, we did not modify the dynamics $p_{\phi}(\mathbf{c}' \mid \mathbf{c}, \mathbf{a})$, that is, the dynamics continued to make predictions using the codebook encoding $\mathbf{c}$ and did not use the one-hot or label encodings.
This is because when we replaced the codebook encoding with either one-hot or label encodings, this led to the
training curves (environment step vs episode return) flat-lining and unable to solve the task.
This suggests that our codebook encoding is needed in our self-supervised world model setup.
Nevertheless, we evaluated the performance when changing the encoding for the other components.

We evaluated the following experiment configurations:
\textbf{Codes (purple)}: All components used codes: dynamics $p_{\phi}(\mathbf{c}' \mid \mathbf{c}, \mathbf{a})$, reward $R_{\xi}(\mathbf{c}, \mathbf{a})$, critic $Q_{\psi}(\mathbf{c}, \mathbf{a})$ and prior policy $\pi_{\eta}(\mathbf{c})$.
\textbf{Label (blue)}: Dynamics model used codes $p_{\phi}(\mathbf{c}' \mid \mathbf{c}, \mathbf{a})$ whilst reward $R_{\xi}(\mathbf{e}_{\text{label}}, \mathbf{a})$, critic $Q_{\psi}(\mathbf{e}_{\text{label}}, \mathbf{a})$ and prior policy $\pi_{\eta}(\mathbf{e}_{\text{label}})$ used labels $\mathbf{e}_{\text{label}}$ obtained from the code's index $i$ in the codebook.
\textbf{One-hot (red)}: Dynamics model used codes $p_{\phi}(\mathbf{c}' \mid \mathbf{c}, \mathbf{a})$ whilst reward $R_{\xi}(\mathbf{e}_{\text{one-hot}}, \mathbf{a})$, critic $Q_{\psi}(\mathbf{e}_{\text{one-hot}}, \mathbf{a})$ and prior policy $\pi_{\eta}(\mathbf{e}_{\text{one-hot}})$, used the one-hot $\mathbf{e}_{\text{one-hot}}$ representation of the label encoding.
}

\rebuttal{
The label encoding (blue) struggles to learn in the Humanoid Walk task and is often less sample efficient than the alternative encodings. This is likely because the label encoding is not expressive enough to model the multi-dimensional ordinal structure of our codebook. Let us provide intuition via a simple example. Our codebook has $b=2$ channels, so two different codes may take the form $e_{\text{codes}}(A)=[0.5, -0.5]$ and $e_{\text{codes}}(B)=[0, 0.5]$. As a result, our codebook encoding  can model ordinal structure in both of its channels, \ie, $e_{\text{codes}}(A)_1>e_{\text{codes}}(B)_1$ whilst $e_{\text{codes}}(A)_2 < e_{\text{codes}}(B)_2$. The corresponding label encoding would encode this as $e_{\text{label}}(A)=1$ and $e_{\text{label}}(B)=2$, which incorrectly implies that $B>A$. In short, the label encoding cannot model the multi-dimensional ordinal structure of the codebook $\mathcal{C}$.
}
In contrast, the one-hot encoding (red) matches the codebook encoding in terms of sample efficiency in all tasks
except Humanoid Walk.
However, the one-hot encoding introduces an extremely large input dimension for the reward, value and policy networks,
and this significantly slows down training.
\rebuttal{See \cref{sec:preliminaries} for further details on why this is the case.}
\begin{figure}[b]
\vskip -0.1in
\begin{center}
\includegraphics[width=0.24\textwidth]{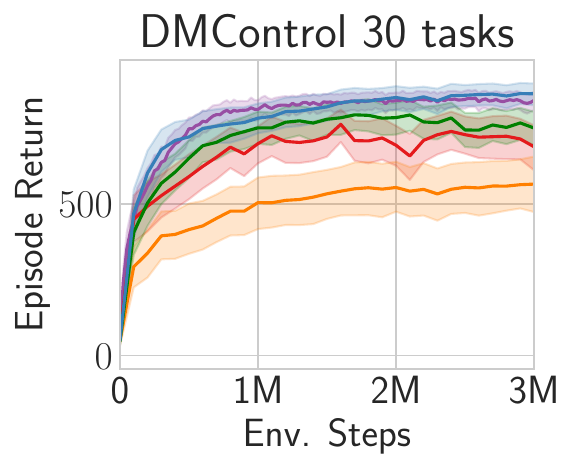}
\includegraphics[width=0.24\textwidth]{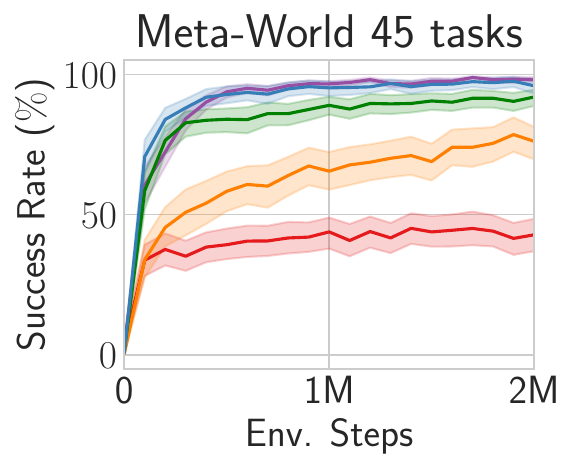}
\includegraphics[width=0.4\textwidth]{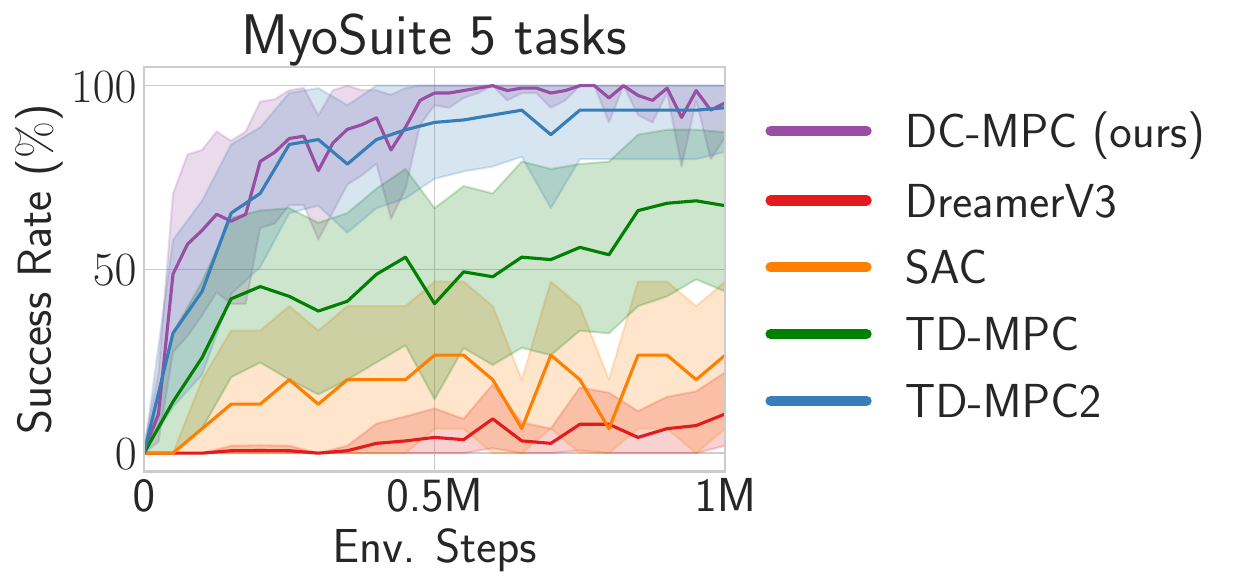}
\caption{\rebuttal{\textbf{Aggregate training curves in DMControl, Meta-World, \& MyoSuite} \our generally matches TD-MPC2 whilst outperforming DreamerV3, SAC and TD-MPC across all tasks. We plot the mean (solid line) and the $95\%$ confidence intervals (shaded) across 3 seeds per task.}}
\label{fig:dmc_mw_rliable}
\end{center}
\vskip -0.2in
\end{figure}

\subsection{Performance of \our}
\rebuttal{
In \cref{fig:dmc_mw_rliable,fig:dmc_rliable,fig:metaworld_rliable,fig:myosuite_rliable}, we compare the aggregate performance of \our against
TD-MPC2, DreamerV3, TD-MPC, and SAC, in 30 DMControl, 45 Meta-World, and 5 MyoSuite tasks respectively, with 3 seeds per task.}
Some tasks in DMControl are particularly high-dimensional. For instance, the observation space of the Dog tasks is $\mathcal{O} \in \R^{223}$ and the action space is $\mathcal{A} \in \R^{38}$, and for Humanoid, the observation space is $\mathcal{O} \in \R^{67}$ and the action space $\mathcal{A} \in \R^{24}$.
\cref{fig:high-dim-locomotion} shows that \our excels in the high dimensional Dog and Humanoid environments when compared to the baselines. We hypothesize that our discretized representations are particularly beneficial for simplifying learning the transition dynamics in high-dimensional spaces, making \our highly sample efficient in these tasks.
Similarly, we find that \our outperforms DreamerV3 in simulated manipulation tasks in the Meta-World task suite (\cref{fig:dmc_mw_rliable,fig:metaworld_grid,fig:metaworld_rliable}).
We also see that \our generally matches the performance of TD-MPC2.
Comparing the results at a global level (\cref{fig:dmc_mw_rliable}), we can find that our proposed method performs well across all benchmarks.%

\begin{figure}[t]
  \centering
  \begin{subfigure}{0.47\textwidth}
    \centering
    \includegraphics[width=\linewidth]{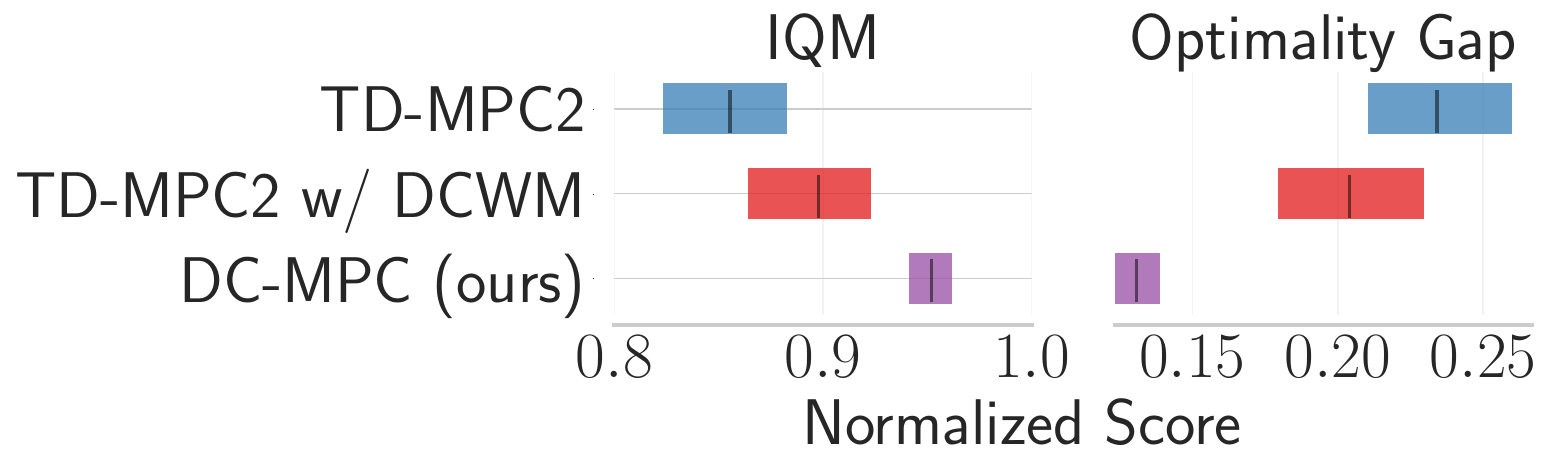}
    \caption{\textbf{Aggregate statistics at 1M environment steps}}
  \end{subfigure}%
  \hfill
  \begin{subfigure}{0.52\textwidth}
    \centering
    \includegraphics[width=0.8\linewidth]{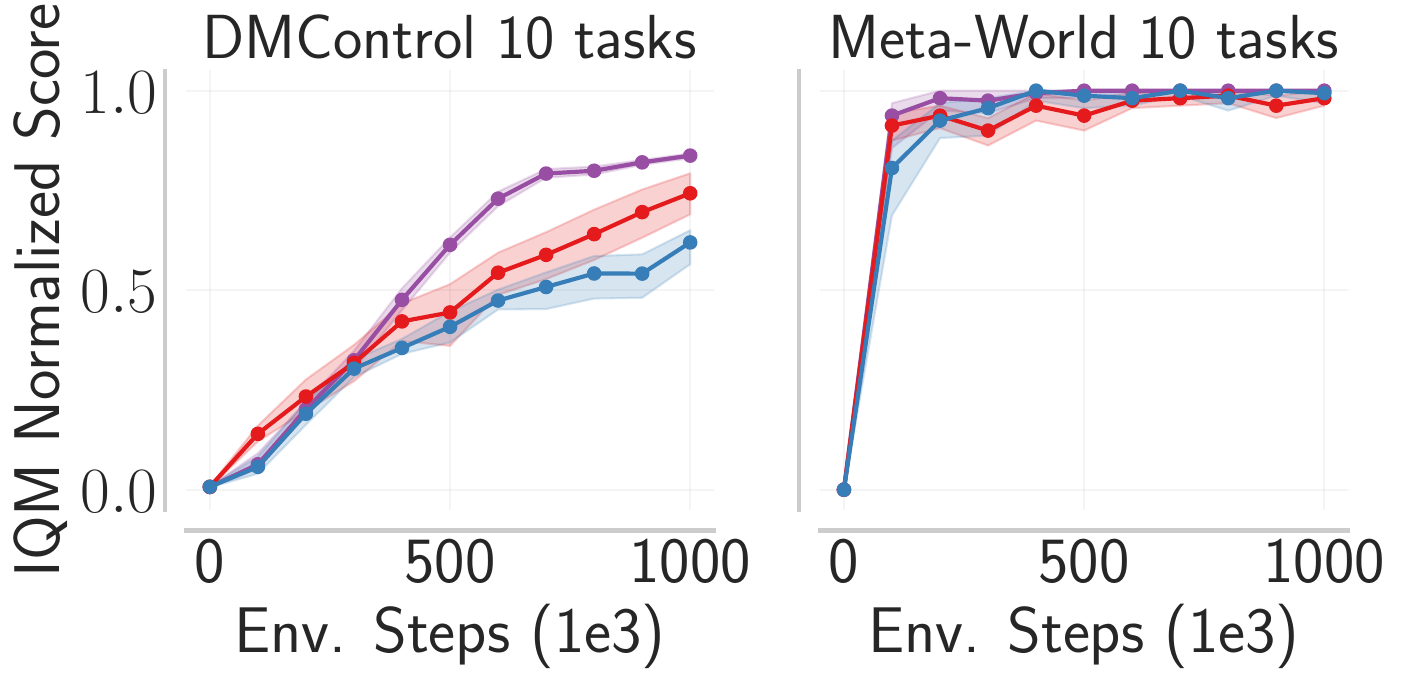}
    \caption{\textbf{Training curves}}
  \end{subfigure}
  \caption{\textbf{TD-MPC2 with \ourM} Adding \our's discrete and stochastic latent space to TD-MPC2 improves performance. See \cref{app:tdmpc2_with_dcwm,app:full_results} for more details.}
  \label{fig:tdmpc2_with_dcwm_rliable}
\vskip -0.2in
\end{figure}

It is important to note that TD-MPC2 has multiple algorithmic differences to \our which means that a straight-up comparison between them is not only affected by
the latent space design.
For example, it {\em (i)} uses soft-actor critic (SAC) to learn the prior policy (helping with exploration in
sparse reward tasks), {\em (ii)} learns the value function jointly with the world model, and
{\em (iii)} the reward and value functions are formulated using discrete
regression in a $\mathrm{log}\text{-transformed}$ space.
In \cref{fig:tdmpc2_with_dcwm_rliable}, we show that incorporating \ourM's stochastic and discrete codebook latent space into TD-MPC2 (red) offers
improvements over vanilla TD-MPC2.
See \cref{app:tdmpc2_with_dcwm,app:full_results} for more details on these experiments and \cref{app:decoder-ablation} where we tried the same experiments with DreamerV3.
DreamerV3's performance is poor in the harder tasks so we do not see any benefit from using \ourM. However, we identify that its poor performance
stems from using observation reconstruction. %

\paragraph{Further Experiments}
In \cref{app:codebook-ablation,app:latent-space-ablation}, we evaluate \our's sensitivity to codebook size $|\mathcal{C}|$ and latent dimension $d$, respectively,
in \cref{app:latent-space-ablation}, we show that stochastic continuous latent spaces do not appear to offer the same benefits as stochastic discrete latent spaces,
in \cref{app:vq-vae-ablation}, we ablate FSQ and show that it either matches or outperforms vector quantization (VQ) whilst being simpler,
and in \cref{app:double-q-ablation}, we show the benefit of using REDQ's ensemble of critics vs the standard double Q approach.\looseness-1

\section{Conclusion}
\label{sec:conclusion}
We have presented \our, a world model that learns a discrete and stochastic latent space using codebook encodings and
a cross-entropy based self-supervised loss for model-based RL.
\our demonstrates strong performance in continuous control tasks, including Meta-World and the complex DMControl Humanoid and Dog tasks, where it exceeds or matches the performance of SOTA baselines.
Our results indicate that using straight-through Gumbel-softmax sampling when making multi-step dynamics predictions is beneficial for
world model learning, both in \our and our experiments where we modified TD-MPC2's latent space.
In summary, we have demonstrated the benefit of a discrete latent space with codebook encodings over a standard continuous latent embedding or classical discrete spaces such as label and one-hot encodings.
These findings open up a new interesting avenue for future research into discrete embeddings for world models.\looseness-1

\paragraph{Limitations and Future Work}
As our goal was to evaluate latent space design, we did not prioritize making \our run with a single set of hyperparameters and
we tuned the noise schedule and the $N\text{-step}$ return for some tasks.
In future work, it would be interesting to make \our robust to hyperparameters.
For example, it would be interesting to model the epistemic uncertainty associated with the latent transition dynamics -- arising from learning from limited data --
and using it to equip \our with a more principled exploration mechanism like
\citet{chuaDeepReinforcementLearning2018,scannell2024functionspace,daxbergerLaplace2021} and remove the task-specific noise schedules.
It would also be interesting to investigate if our results hold for different world model
backbones \citep{dengFacingWorldModel2023,nvidia2025cosmosworldfoundationmodel},
such as Transformers
\citep{vaswaniAttentionAllYou2017,robineTransformerbasedWorldModels2022,zhangSTORMEfficientStochastic2023,micheliTransformersAreSampleEfficient2022,barNavigationWorldModels2024}
and diffusion models \citep{hoDenoisingDiffusionProbabilistic2020,alonsoDiffusionWorldModeling2024}.
Finally, it would be interesting to investigate how well \our scales
\citep{kaplanScalingLawsNeural2020,henighanScalingLawsAutoregressive2020,hoffmannTrainingComputeOptimalLarge2022}
and if it is an effective setup for generalist (\ie multi-embodiment) world modeling \citep{reed2022generalist,ZhaoLeveraging2024}.\looseness-3

\subsubsection*{Acknowledgments}
Aidan Scannell and Kalle Kujanpää were supported by the Research Council of Finland from the Flagship program: Finnish Center for Artificial Intelligence (FCAI).
Arno Solin and Yi Zhao acknowledge funding from the Research Council of Finland (grant ids 339730 and 357301, respectively) and Mohammadreza Nakhaei acknowledges funding from Business Finland (BIOND4.0 -- Data Driven Control for Bioprocesses).
Kevin Sebastian Luck is supported by the project \textit{TeNet: Text-to-Network for Fast and Energy-Efficient Robot Control} with file number NGF.1609.241.015 of the research programme National Growth Fund AiNed XS Europe 24-2 which is financed by the Dutch Research Council (NWO). 
We acknowledge CSC -- IT Center for Science, Finland, for awarding this project access to the LUMI supercomputer, owned by the EuroHPC Joint Undertaking, hosted by CSC (Finland) and the LUMI consortium through CSC.
We acknowledge the computational resources provided by the Aalto Science-IT project.

\newpage

\bibliographystyle{iclr2025_conference}

\clearpage
\appendix
\section*{Appendices}

This appendix is organized as follows.
In \cref{app:method} we provide further details on our method.
\cref{app:full_results} provides further experimental results, including  evaluating \our's sensitivity to the codebook size in \cref{app:codebook-ablation},
its sensitivity to latent dimension in \cref{app:latent-dim-ablation},
further details on the latent space ablation in \cref{app:latent-space-ablation},
a comparison of \our using VQ instead of FSQ in \cref{app:vq-vae-ablation},
a comparison of \our's ensemble REDQ critic approach vs the standard double Q approach in \cref{app:double-q-ablation},
full DeepMind control suite results in \cref{app:full_dmc},
Meta-World results in \cref{app:full_metaworld},
MyoSuite results in \cref{app:myosuite},
evaluation of DreamerV3 using \our's latent space in \cref{app:decoder-ablation}
and an evaluation of TD-MPC2 using \our's latent space in  \cref{app:tdmpc2_with_dcwm}.
In \cref{app:implementation}, we provide further implementation details, including default hyperparameters, hardware, \etc
In \cref{app:baselines}, we provide further details of the baselines and
in \cref{app:tasks} we detail the different DeepMind control, Meta-World and MyoSuite tasks used throughout the paper.

\medskip

\begin{sc}
 \startcontents[appendices]
 \printcontents[appendices]{l}{1}{}%
\end{sc}

\vfill
\begin{center}
    --appendices continue on next page--
\end{center}

\clearpage
\section{Method details}
\label{app:method}

\cref{alg:world_model_alg} outlines \our's training procedure.
\begin{algorithm}[h]
   \caption{\our's training}
   \label{alg:world_model_alg}
   \renewcommand{\algorithmiccomment}[1]{\hfill\textcolor{gray}{\(\triangleright\) #1}}
\begin{algorithmic}
   \STATE {\bfseries Input:} Encoder $e_{\theta}$, dynamics $d_{\phi}$, reward $R_{\xi}$, critics $\{q_{\psi_{i}}\}_{i=1}^{N_{q}}$, policy $\pi_{\eta}$, learning rate $\alpha$, target network update rate $\tau$, episode length $T$, replay buffer $\mathcal{D}=\{\}$
   \FOR{$1:N_{\text{random episodes}}$}
    \STATE $\mathcal{D} \leftarrow \mathcal{D} \cup \{\vo_{t}, \va_{t}, \vo_{t+1}, r_{t}\}^{T}_{t=0}$ \COMMENT{Collect data using random policy}
    \ENDFOR
   \FOR{$1:N_{\text{episodes}}$}
    \STATE $\mathcal{D} \leftarrow \mathcal{D} \cup \{\vo_{t}, \va_{t}, \vo_{t+1}, r_{t}\}^{T}_{t=0}$ \COMMENT{Collect data using \our}
    \FOR{$i=1$ {\bfseries to} $T$}
        \STATE $[\theta, \phi, \xi] \leftarrow [\theta, \phi, \xi] + \alpha \nabla \left( \mathcal{L}(\theta, \phi, \xi; \mathcal{D}) \right)$  \COMMENT{Update world model, \cref{eq:world-model-loss}}
        \STATE $\psi \leftarrow \psi + \alpha \nabla \left( \mathcal{L}_{q}(\psi; \mathcal{D}) \right)$ \COMMENT{Update critic, \cref{eq:value-loss}}
        \IF{$i$ \% 2 == 0}
          \STATE $\eta \leftarrow \eta + \alpha \nabla \left( \mathcal{L}_{\pi}(\eta; \mathcal{D}) \right)$  \COMMENT{Update actor less frequently than critic, \cref{eq:policy-loss}}
        \ENDIF
        \STATE $ [\bar{\psi}, \bar{\eta}] \leftarrow (1-\tau)  [\bar{\psi}, \bar{\eta}] + \tau [{\psi}, \eta]$ \COMMENT{Update target networks}
    \ENDFOR
   \ENDFOR
\end{algorithmic}
\end{algorithm}

\cref{alg:mppi_alg} outlines how we perform trajectory optimization using MPPI \citep{williams2015model}, closely following the formulation of MPPI by \citet{hansenTemporalDifferenceLearning2022}, with two key modifications.
First, during each rollout, we use the expected next latent state, \ie a weighted sum over the codes in the codebook.
Note that this contrasts our world model training where we sample from the transition dynamics $p(\vc_{h+1} | \vc_h, \vc_h)$.
This approach reduces the variance in state transitions, which results in more stable trajectory evaluations.
Second, we do not add noise sampled from the standard deviation $\bm\sigma^2_{0}$  returned from MPPI.
Instead, we promote exploration by adding noise sampled from a separate noise schedule.%
This method, inspired by TD3 \citep{fujimotoAddressingFunctionApproximation2018}, strikes a better balance between exploration and exploitation, leading to more stable training performance.

It is worth noting that MPPI resembles the CEM-based planner in \citet{chuaDeepReinforcementLearning2018}, however, instead of simply fitting a
Gaussian to the top $K$ action samples at each iteration,
MPPI uses weighted importance sampling, which weights \textbf{all} samples by their empirical return estimates.
However, we follow \citet{hansenTemporalDifferenceLearning2022} and use a hybrid approach, which selects the top $K$ action samples
(like CEM) but then use weighted importance sampling (like MPPI).
At each iteration, we calculate the mean and variance of the action trajectory as follows,
\begin{align}
  \bm\mu_{0:H} =\mathrm{fit\_mean}\Big(\big\{ \big(\va_{0:H}^{(i)}, \Phi^{(i)} \big) \big\}_{i=0}^{K} \Big) &=
   \sum_{i=1}^{K}  \underbrace{\frac{\Omega^{(i)} }{\sum_{j=1}^{K} \Omega^{(j)}}}_{\text{importance weight}} \va_{0:H}^{(i)} \\
  \bm\sigma^{2}_{0:H} = \mathrm{fit\_var}\Big(\big\{ \big(\va_{0:H}^{(i)}, \Phi^{(i)} \big) \big\}_{i=0}^{K} \Big) &=
   \frac{\sum_{i=1}^{K} \Omega^{(i)} \left(\va_{0:H}^{(i)} - \bm\mu_{0:H} \right)^{2}}{\sum_{i=1}^{K}\Omega^{(i)}}
\end{align}
where $\Omega^{(i)}$ is the exponentiated normalized empirical return estimate given by
{$\Omega^{(i)} = \mathrm{exp} \left(\tau_{\text{MPPI}}  \left(\Phi^{(i)} - \mathrm{max} \left(\{\Phi^{(0)}, \ldots, \Phi^{(N_p + N_{\pi})} \} \right) \right) \right)$}.
Note that $\tau_{\text{MPPI}}$ is the (inverse) temperature parameter and $\Phi^{(i)}$ denotes the return estimate for the $i^{\text{th}}$ action trajectory $\va_{0:H}^{(i)}$.
After $J$ (default $6$) iterations, we sample one of the top $K$ action sequences $\{\va_{0:H}^{(i)}\}_{i \in \{i^{*}_{1},\ldots,i^{*}_{K}\}}$
where each action sequence is weighted by its empirical return estimate $\{\Omega^{(i)} \}_{i \in \{i^{*}_{1},\ldots,i^{*}_{K}\}}$.
We then apply the first action $\va_{0}^{(i^{*})}$ in the environment.

\vfill
\begin{center}
    --appendices continue on next page--
\end{center}
\clearpage

\begin{algorithm}[h]
   \caption{\our's inference (modified MPPI)}
   \label{alg:mppi_alg}
   \renewcommand{\algorithmiccomment}[1]{\hfill\textcolor{gray}{\(\triangleright\) #1}}
\begin{algorithmic}
   \STATE {\bfseries Input:} current observation $\vo$, planning horizon $H$, iterations $J$, population size $N_{p}$, prior population size $N_{\pi}$, number of elites $K$, exploration noise std $\sigma_{\text{noise}}$
    \STATE $\vc_0 \leftarrow e_{\theta}(\vo)$ \COMMENT{Encode observation into discrete code}
    \STATE Initialize $\vmu_{0:H}^0$, $(\bm\sigma^{2}_{0:H})^0$ with the solution from the last time step shifted by one.
   	\FOR{each iteration $j=1,\ldots,J$}
    \STATE Sample $N_{p}$ action trajectories of length $H$ from $\{a_{h} \sim \mathcal{N}(\vmu^{j-1}_{h}, (\bm\sigma^{2}_{h})^{j-1})\}_{h=0}^{H}$ \COMMENT{Sample action candidates}
    \STATE Sample $N_{\pi}$ action trajectories of length $H$ using $\pi_{\eta}$ and $d_{\phi}$ \COMMENT{Prior policy samples}
    \FOR{all $N_{p} + N_\pi$ action sequences $\left\{\tau^{(i)} = \left(\va_0^{(i)}, \dots, \va_H^{(i)}\right)\right\}_{i=1}^{N_{p}+N_{\pi}}$} \COMMENT{Trajectory evaluation}
    \STATE $\Phi^{(i)} \leftarrow 0$
    \FOR{step $h=0,\ldots,H-1$}
      \STATE $\Phi^{(i)} \leftarrow \Phi^{(i)} + \gamma^{h} R_{\xi}(\hat{\vc}_{h}, \va_{h}^{(i)})$ \COMMENT{Compute immediate reward}
      \STATE $\hat{\vc}_{h+1} = \sum_{k=1}^{|\mathcal{C}|} \Pr(\hat{\vc}_{h+1}=\vc^{(k)} \mid \hat{\vc}_{h}, \va_{h}^{(i)}) \vc^{(k)}$ \COMMENT{Compute next state}
    \ENDFOR
    \STATE $\Phi^{(i)} \leftarrow \Phi^{(i)} + \gamma^{H} \frac{1}{N_q}\sum_{k=1}^{N_{q}}q_{\psi_k}(\vc_{H},\va_{H}^{(i)})$ \COMMENT{Bootstrap with ensemble of Q-functions}
    \ENDFOR
    \STATE $\Phi^{(i^*_{1})}, \ldots, \Phi^{(i^*_{K})} = \mathrm{topk}(\{ \Phi^{(0)}, \ldots, \Phi^{(N_{p}+N_{\pi})} \})$ \COMMENT{Get top-$K$ elite scores}
    \STATE $\bm\mu_{0:H} \leftarrow \mathrm{fit\_mean}\Big(\big\{\big( \va_{0:H}^{(i)}, \Phi^{(i)} \big)\big\}_{i \in \{i^{*}_{1}, \ldots, i^{*}_{K}\}} \Big)$ \COMMENT{Update mean of action dist.}
    \STATE $\bm\sigma^{2}_{0:H} \leftarrow \mathrm{fit\_var}\Big(\big\{\big( \va_{0:H}^{(i)}, \Phi^{(i)} \big)\big\}_{i \in \{i^{*}_{1}, \ldots, i^{*}_{K}\}} \Big)$ \COMMENT{Update variance of action dist.}
   \ENDFOR
    \STATE $i^{*} \sim \mathrm{Categorical}\left( \mathrm{softmax}(\{\Phi^{(i^{*}_{1})}, \ldots, \Phi^{(i^{*}_{K})}\}) \right)$ \COMMENT{Sample action index according to scores}
    \STATE\textbf{return}~$\va_0^{(i^{*})} + \epsilon \quad \text{with} \quad  \epsilon \sim \mathcal{N}(0,\sigma^2_{\text{noise}})$ \COMMENT{Final output with exploration noise}
\end{algorithmic}
\end{algorithm}

\vfill
\begin{center}
    --appendices continue on next page--
\end{center}
\clearpage
\section{Further Results}
\label{app:full_results}
In this section, we include further results and ablations.

\paragraph{Aggregate metrics}
In \cref{fig:dmc_rliable,fig:metaworld_rliable,fig:myosuite_rliable}, we compare the aggregate performance of \our against TD-MPC, TD-MPC2,
DreamerV3, and SAC, in 30 DMControl tasks, 45 Meta-World tasks, and 5 MyoSuite tasks respectively, with 3 seeds per task.
Following \citet{agarwalDeepReinforcementLearning2021}, we report the median, interquartile mean (IQM), mean,
and optimality gap at 1M environment steps, with error bars representing $95\%$ stratified bootstrap confidence intervals.
For DMControl, we use min-max normalization as the maximum possible return in an episode is $1000$ whilst the minimum is $0$, \ie
$\text{Normalized Return} = \text{Return} / (1000-0)$.
For Meta-World, we report the success rate which does not require normalization as it is already between $0$ and $1$.

In \cref{fig:tdmpc2_with_dcwm_rliable,fig:latent-space-comparison} we report aggregate metrics over 10 DMControl and 10 Meta-World tasks.
The tasks are as follows:
\begin{itemize}
  \item \textbf{DMControl 10}: Acrobot Swingup, Dog Run, Dog Walk, Dog Stand, Dog Trot, Humanoid Stand, Humanoid Walk, Humanoid Run, Reacher Hard, Walker Walk.
  \item \textbf{Meta-World 10}: Button Press, Door Open, Drawer Close, Drawer Open, Peg Insert Side, Pick Place, Push, Reach, Window Open, Window Close.
\end{itemize}

\vfill
\begin{center}
    --appendices continue on next page--
\end{center}

\clearpage
\subsection{Sensitivity to Codebook Size $|\mathcal{C}|$}
\label{app:codebook-ablation}

\begin{figure}[h]
	\begin{center}
			\centerline{\includegraphics[width=\textwidth]{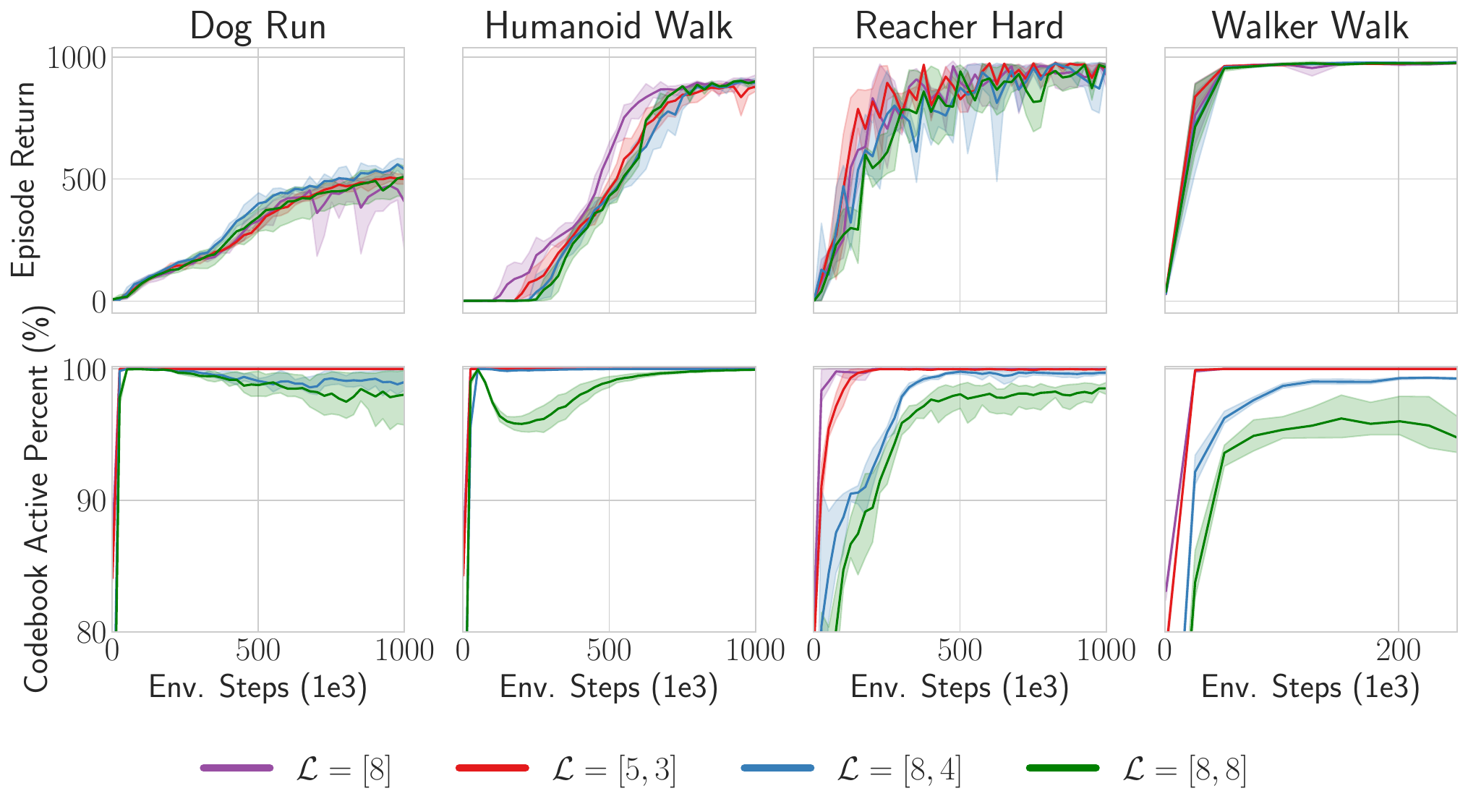}}
			\caption{\textbf{Sensitivity to codebook size} We compare how the codebook size affects the performance of \our (top) and the percentage of the codebook that is active during training (bottom). In general, smaller codebooks become fully active faster than larger codebooks. We plot the mean and the $95\%$ confidence intervals (shaded) across 3 random seeds for all environments.}
			\label{fig:codebook-size-ablation}
		\end{center}
\end{figure}

In this section, we evaluate how the size of the codebook $|\mathcal{C}|$ influences training.
We indirectly configure different codebook sizes via the FSQ levels $\mathcal{L} = \{L_{1},\ldots,L_{b} \}$ hyperparameter.
This is because the codebook size is given by $|\mathcal{C}| = \prod_{i=1}^{b} L_{i}$. The top row of \cref{fig:codebook-size-ablation} compares the training curves for different codebook sizes. The algorithm's performance is not particularly sensitive to the codebook size. A codebook that is too large can result in slower learning. The best codebook size varies between environments.

Given that a codebook has a particular size, we can gain insights into how quickly \our's encoder starts to activate all of the codebook. The connection between the codebook size and the activeness of the codebook is intuitive: the bottom row of \cref{fig:codebook-size-ablation} shows that the smaller the codebook, the larger the active proportion.

\vfill
\begin{center}
    --appendices continue on next page--
\end{center}

\clearpage

\subsection{Sensitivity to Latent Dimension $d$}
\label{app:latent-dim-ablation}
\begin{figure}[h]
	\begin{center}
			\centerline{\includegraphics[width=\textwidth]{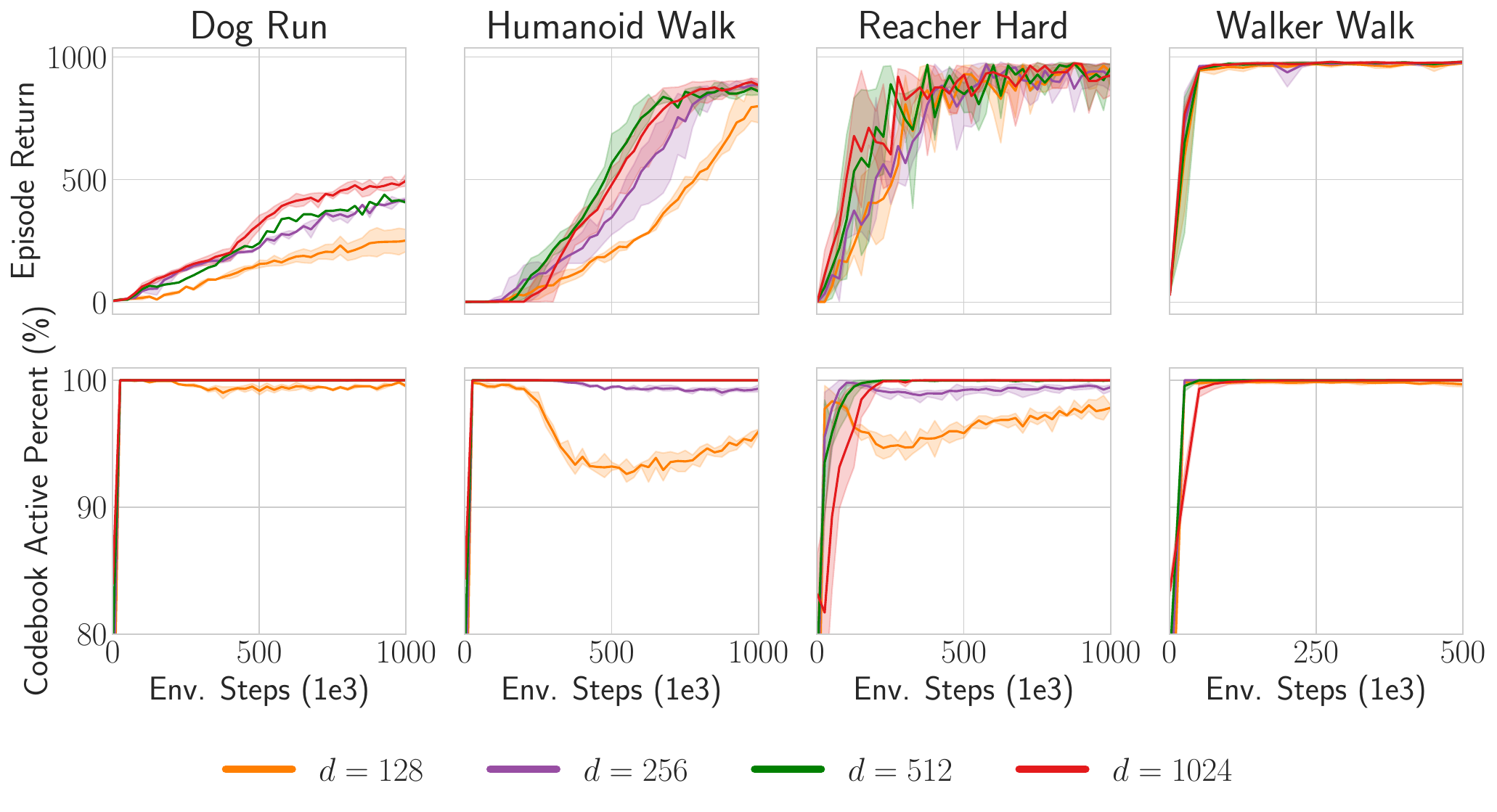}}
			\caption{\textbf{Sensitivity to latent dim $d$} We compare how the latent dimension $d$ affects the performance of \our (top) and the percentage of the codebook that is active during training (bottom). In general, our algorithm is robust to the latent dimension of the representation, although in more difficult environments, such as Humanoid Walk, a $d$ too small can harm the agent's performance. We plot the mean and the $95\%$ confidence intervals (shaded) across 3 random seeds for all environments.}
			\label{fig:latent-dim-ablation}
		\end{center}
\end{figure}

This section investigates how the latent dimension $d$ affects the behavior and performance of \our in four different environments.
In the top row of \cref{fig:latent-dim-ablation}, we see that the performance of our algorithm is robust to the latent dimension $d$, although a latent dimension too small can result in inferior performance, especially in the more difficult environments. The bottom row of \cref{fig:latent-dim-ablation} demonstrates that \our learns to use the complete codebook irrespective of the latent dimension.

\vfill
\begin{center}
    --appendices continue on next page--
\end{center}

\clearpage

\subsection{Ablation of Latent Space}
\label{app:latent-space-ablation}
\begin{figure}[h]
  \centering
  \includegraphics[width=0.9\textwidth]{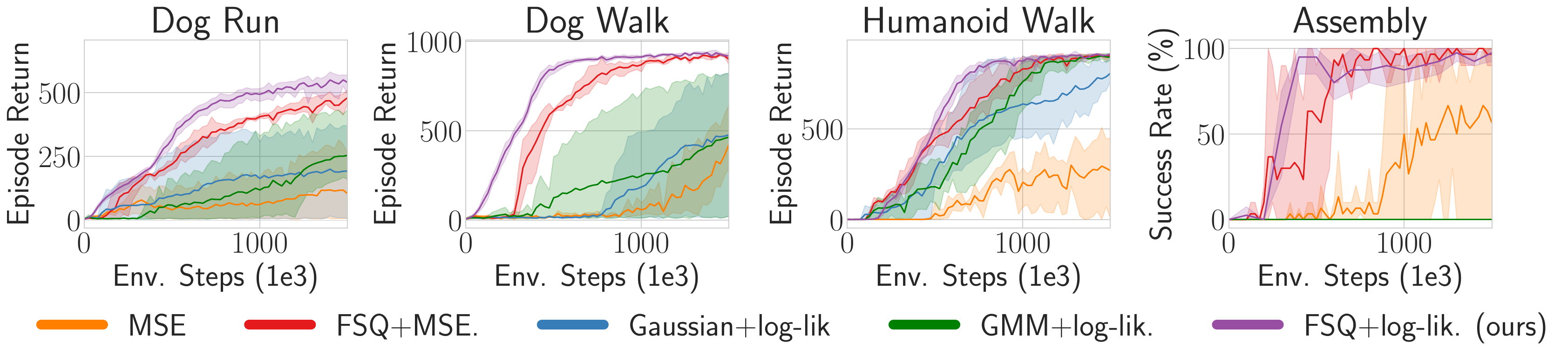}
  \caption{\textbf{Latent space comparison}
    Comparison of different latent space formulations. Continuous and deterministic latent space trained with MSE regression (orange), deterministic and discrete trained with MSE (red), continuous and unimodal Gaussian latent space trained with maximum log-likelihood (blue), continuous and multimodal GMM trained with maximum log-likelihood (green), and discrete trained with classification (purple).
  Discretizing the latent space with FSQ (red) improves sample efficiency and making the dynamics stochastic and training with classification (purple) improves performance further.}
  \label{fig:gaussian-ablation}
  \vskip-0.1in
\end{figure}

In this section, we provide further details on the comparison of different latent spaces experiments
in \cref{sec:latent-space-ablation}.
To validate our method, we test the importance of quantizing the latent space and training the world model
with classification instead of regression.
In \cref{fig:gaussian-ablation}, we compare \our to world models with different latent spaces formulations,
which we now detail.

\paragraph{MSE (orange)}
First, we consider a continuous latent space with deterministic transition dynamics trained by minimizing the
mean squared error between predicted next latent states and target next latent states.

\paragraph{FSQ+MSE (red)}
Next, we consider quantization of the latent space and training based on mean squared error regression.
This experiment allows us to analyze the importance of quantization.

\paragraph{Gaussian+log-lik. (blue)}
To consider stochastic continuous dynamics, we configure the transition dynamics to model a Gaussian distribution
over predictions of the next state.
During training, we sample from the Gaussian distribution using the reparameterization trick.
The world model is then trained to maximize the log-likelihood of the next latent state targets.
This allows us to investigate if modeling stochastic transition dynamics offers benefits when using continuous latent spaces.

\paragraph{GMM+log-lik. (green)}
To consider continuous multimodal transitions, we consider a Gaussian mixture with three components.
During training, we sample a Gaussian from the mixture with the ST Gumbel-softmax trick and then we sample from the
selected Gaussian using the reparameterization trick.
The world model is then trained to maximize the log-likelihood of next latent state targets.

\vfill
\begin{center}
    --appendices continue on next page--
\end{center}

\clearpage

\rebuttal{
\subsection{Ablation of FSQ vs Vector Quantization (VQ)}
\label{app:vq-vae-ablation}
To understand how the choice of using FSQ for discretization contributes to the performance of our algorithm, we tried replacing the FSQ layer with a standard Vector Quantization layer. We evaluated the methods in Walker Walk, Dog Run, Humanoid Walk, and Reacher Hard. We used standard hyperparameters, $\beta=0.25$, and an EMA-updated codebook with a size of $256$ and either $256$ (dog) or $128$ (other tasks) channels per dimension.
We did not change other hyperparameters from \our.
However, we found that to approach the performance of standard FSQ, VQ needs environment-dependent adjusting of the planning procedure. In Humanoid Walk, the performance of FSQ aligns closely with the VQ with a weighted sum over the codes in the codebook for planning (expected code) but significantly outperforms sampled VQ. Conversely, standard sampling is superior in Reacher Hard, which is unsurprising, as the discrete codes in VQ have not been ordered like in FSQ. The necessary environment-specific adjustments for VQ undermine its general applicability compared to FSQ.

\begin{figure}[h]
	\begin{center}
			\centerline{\includegraphics[width=\textwidth]{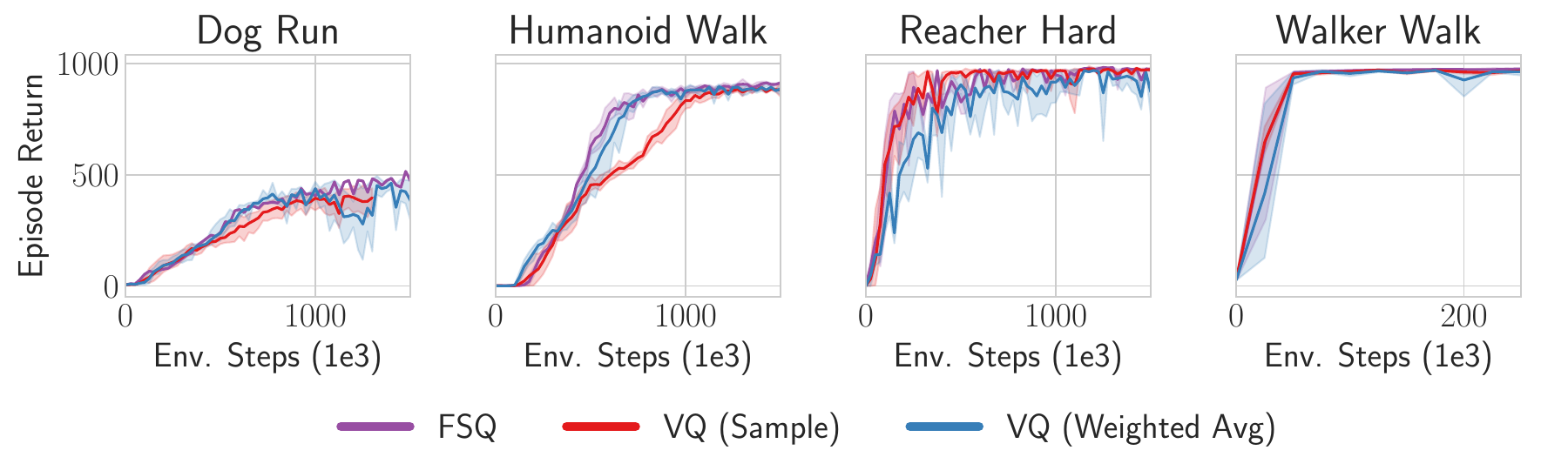}}
			\caption{\textbf{Ablation of FSQ vs VQ} FSQ does not require the extra loss terms required by VQ and it generally performs equal to or better and VQ.}
			\label{fig:vq-vae-ablation}
		\end{center}
\end{figure}
}

\vfill
\begin{center}
    --appendices continue on next page--
\end{center}

\clearpage

\rebuttal{
\subsection{Ablation of REDQ Critic vs Standard Double Q Approach}
\label{app:double-q-ablation}
In this section, we compare the ensemble of Q-functions approach,
used by \our, REDQ \citep{chenRandomizedEnsembledDouble2021} and TD-MPC2 \citep{hansenTDMPC2ScalableRobust2023},
to the standard double Q approach \citep{fujimotoAddressingFunctionApproximation2018}.
In \cref{fig:double-q-ablation}, we evaluate how our default ensemble size of $N_{q}=5$ (purple)
compares with the standard double Q approach, which is obtained by setting the ensemble size
to $N_{q}=2$ (blue).
Note that we always sample two critics so the $N_{q}=2$ result reduces to the standard double Q approach.
\cref{fig:double-q-ablation} shows that \our works fairly well with both approaches but the ensemble approach offers benefits
in the harder Dog Run and Humanoid Walk tasks.

\begin{figure}[h]
	\begin{center}
			\centerline{\includegraphics[width=\textwidth]{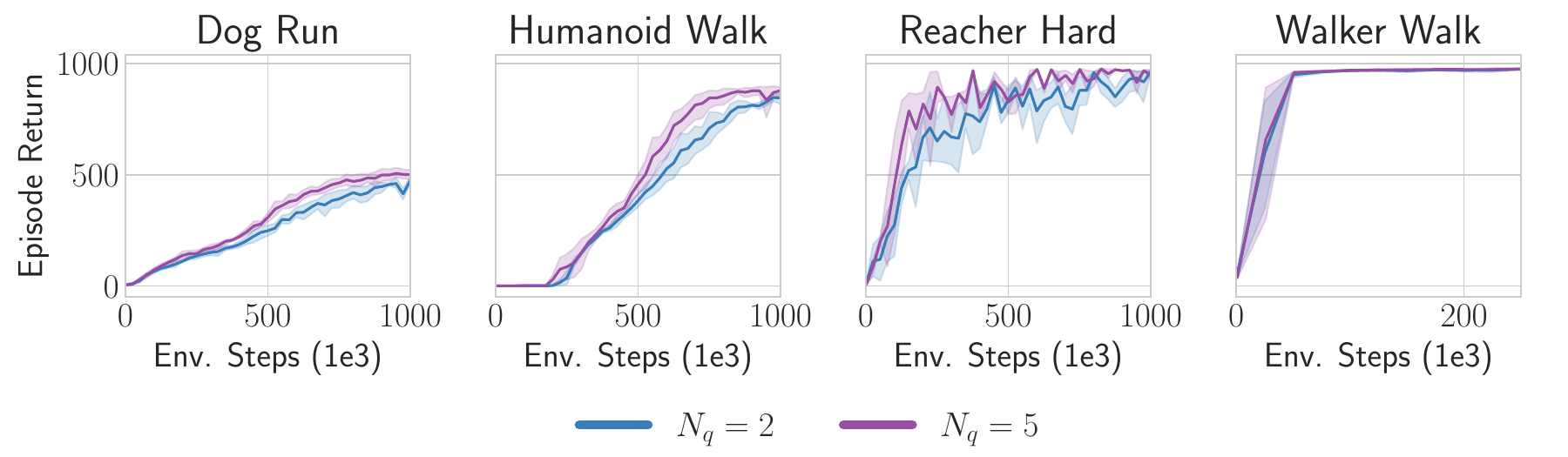}}
			\caption{\textbf{Ablation of REDQ critic vs standard double Q} \our uses a Q ensemble, similar to REDQ, of size $N_{q}=5$ (purple) and sub samples two critics when calculating the mean or minimum Q-value. We compare this approach to the standard double Q approach by setting $N_{q}=2$ (blue) and we see that the ensemble approach offers a slight benefit in the harder Dog Run and Humanoid Walk.}
			\label{fig:double-q-ablation}
		\end{center}
\end{figure}
}

\vfill
\begin{center}
    --appendices continue on next page--
\end{center}

\clearpage

\subsection{DeepMind Control Results}
\label{app:full_dmc}

\begin{figure}[h]
\vskip 0.2in
\begin{center}
\centerline{\includegraphics[width=\textwidth]{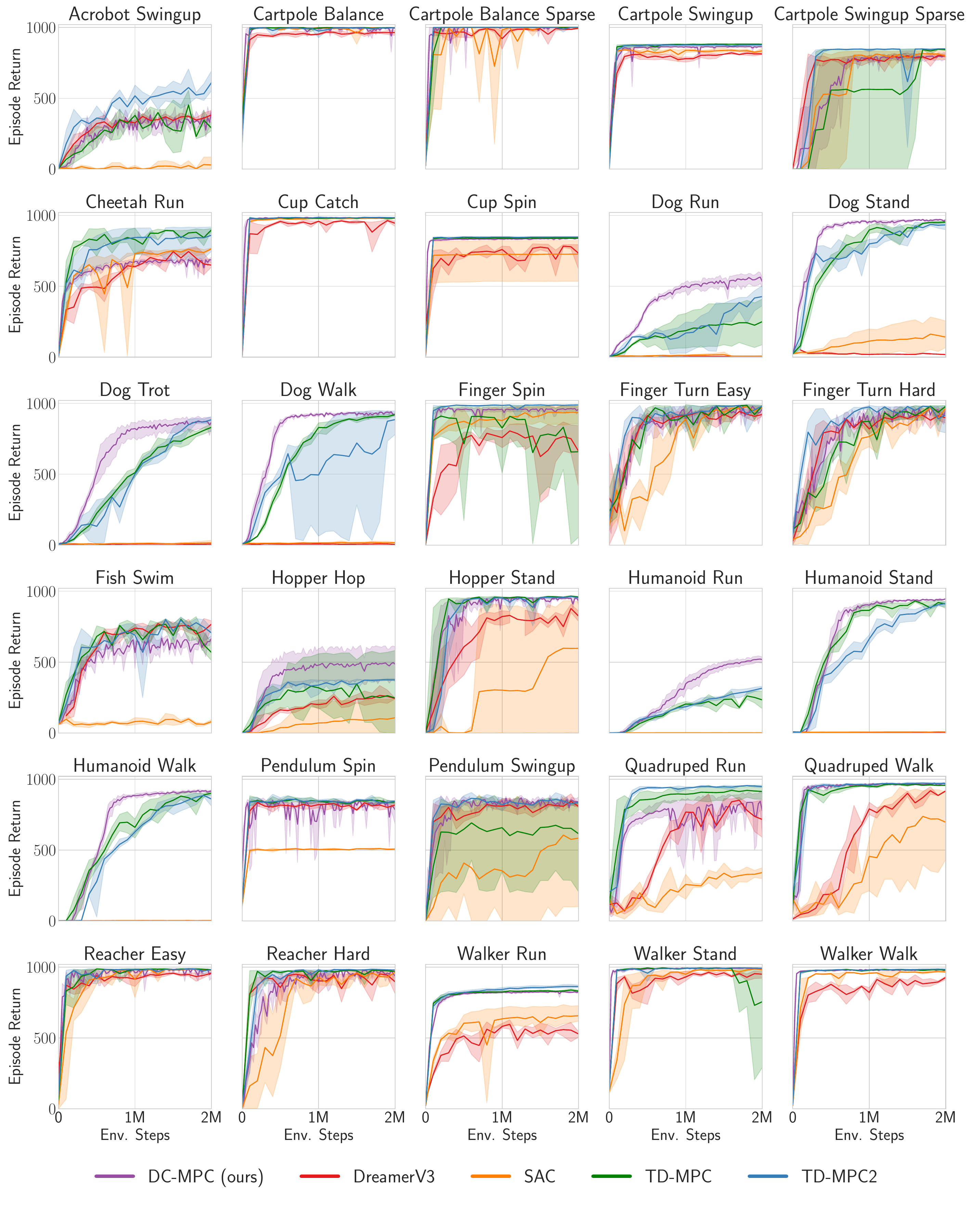}}
\caption{\textbf{DeepMind Control results.} \our performs well across a variety of DMC tasks.
We plot the mean (solid line) and the $95\%$ confidence intervals (shaded) across 5 seeds (\our) or 3 seeds (TD-MPC2/TD-MPC/DreamerV3/SAC), where each seed averages over 10 evaluation episodes.}
\label{fig:dmcontrol_grid}
\end{center}
\vskip -0.2in
\end{figure}

\vfill
\begin{center}
    --appendices continue on next page--
\end{center}
\clearpage

\begin{figure}[h]
\includegraphics[width=\textwidth]{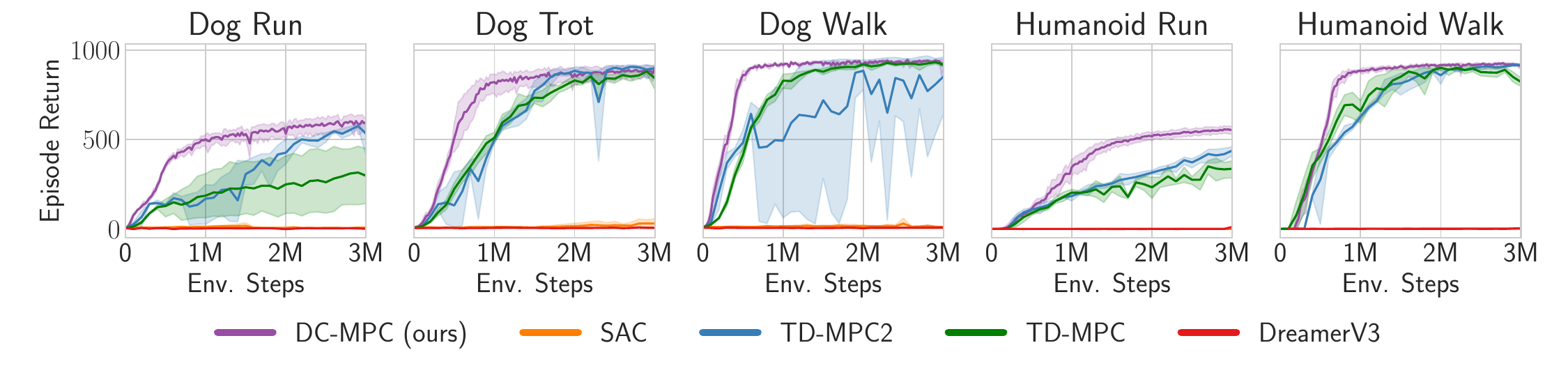}
\caption{\textbf{High-dimensional locomotion} \our (purple) significantly outperforms TD-MPC2 (blue) and DreamerV3 (red) in the complex, high-dimensional locomotion tasks from DMControl.}%
\label{fig:high-dim-locomotion}
\end{figure}

\rebuttal{
\begin{figure}[h]
\begin{center}
\centerline{\includegraphics[width=\textwidth]{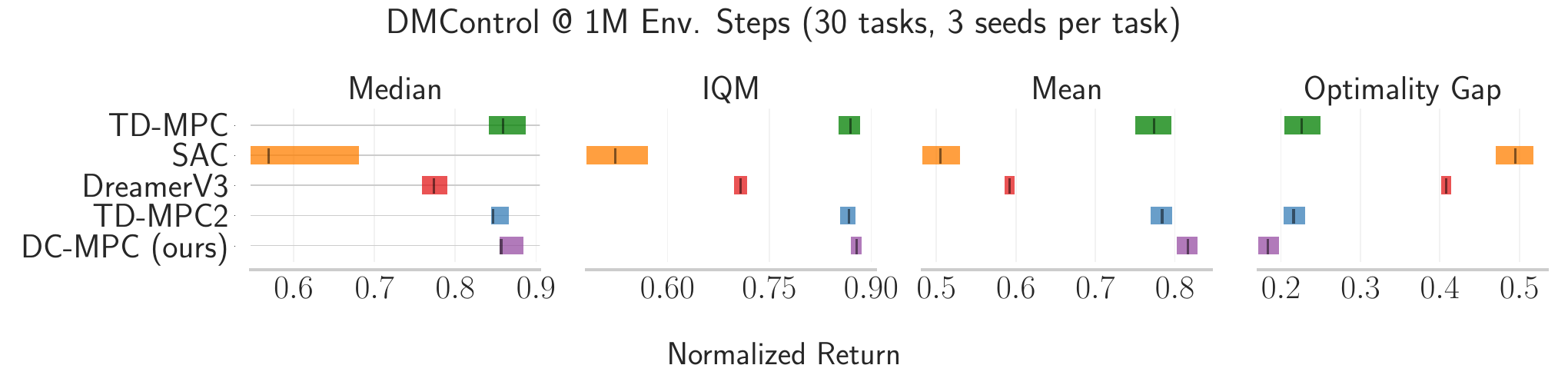}}
\caption{\rebuttal{\textbf{DMControl aggregate results} \our generally outperforms TD-MPC2 and DreamerV3 in DMControl tasks. This is due to \our's strong performance in the hard Dog and Humanoid tasks. Error bars represent $95\%$ stratified bootstrap confidence intervals.
  }
}
\label{fig:dmc_rliable}
\end{center}
\end{figure}
}

\vfill
\begin{center}
    --appendices continue on next page--
\end{center}

\clearpage
\rebuttal{
\subsection{Meta-World Manipulation Results}
\label{app:full_metaworld}

\begin{figure}[h]
\vskip 0.2in
\begin{center}
\centerline{\includegraphics[width=\textwidth]{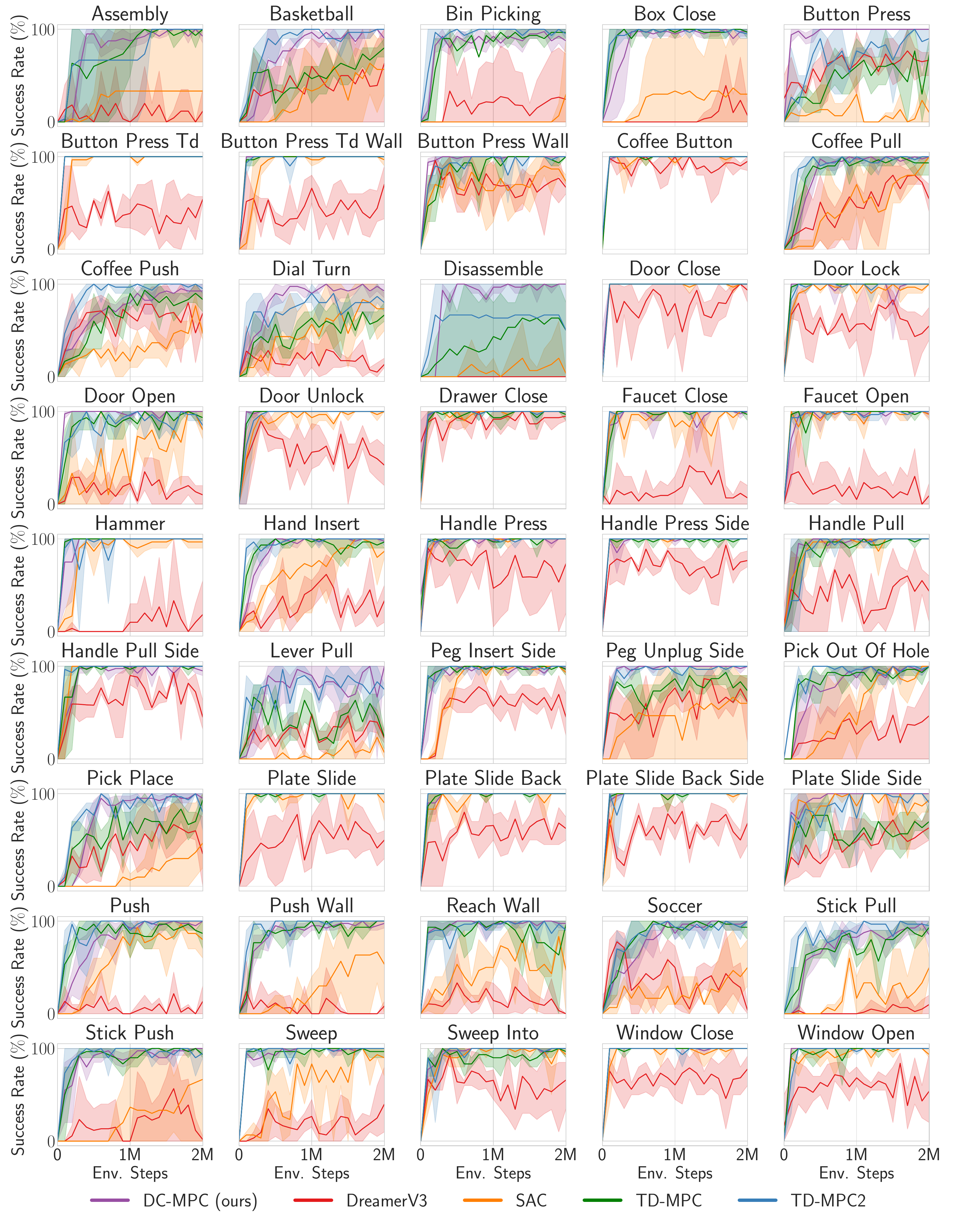}}
\caption{\textbf{Meta-World manipulation results} \our performs well across Meta-World tasks.
We plot the mean (solid line) and the $95\%$ confidence intervals (shaded) across 3 seeds, where each seed averages over 10 evaluation episodes.}
\label{fig:metaworld_grid}
\end{center}
\vskip -0.2in
\end{figure}
}

\vfill
\begin{center}
    --appendices continue on next page--
\end{center}

\clearpage

\rebuttal{
\begin{figure}[h]
\begin{center}
\centerline{\includegraphics[width=\textwidth]{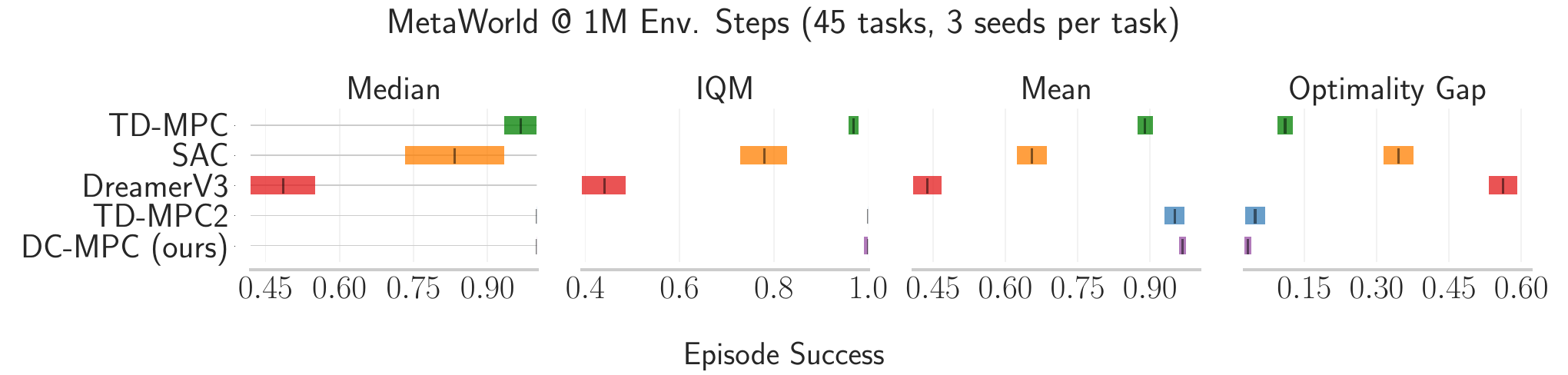}}
\caption{\rebuttal{\textbf{Meta-World results} \our performs well in Meta-World, generally matching TD-MPC2, whilst significantly outperforming DreamerV3 and SAC. Error bars represent $95\%$ stratified bootstrap confidence intervals.
}}
\label{fig:metaworld_rliable}
\end{center}
\vskip -0.3in
\end{figure}
}

\vfill
\begin{center}
    --appendices continue on next page--
\end{center}

\clearpage

\begin{figure}[h]
\vskip 0.2in
\begin{center}
\centerline{\includegraphics[width=\textwidth]{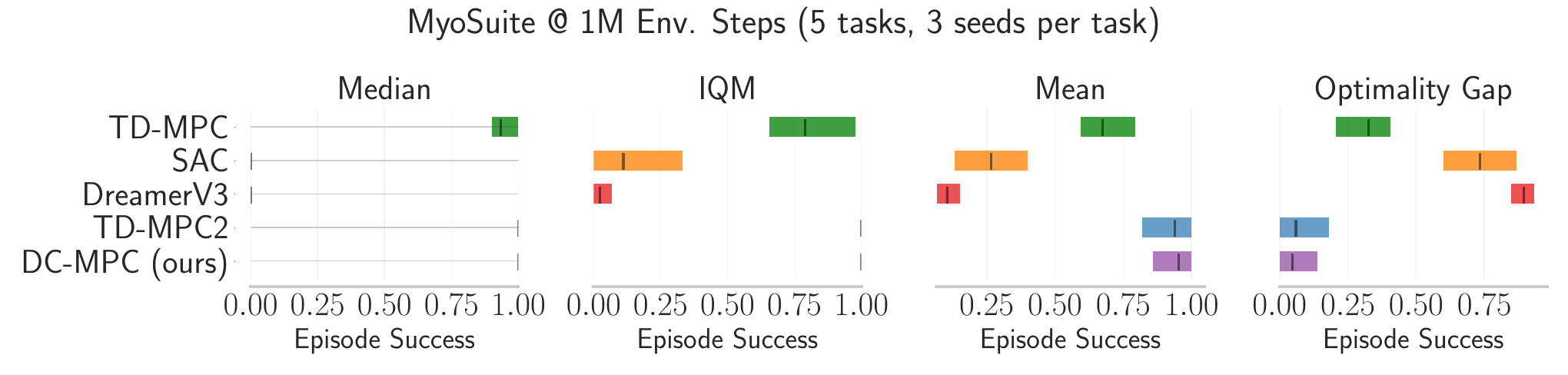}}
\caption{\textbf{MyoSuite results} \our performs similarly to TD-MPC2 in MyoSuite. Error bars represent $95\%$ stratified bootstrap confidence intervals.}
\label{fig:myosuite_rliable}
\end{center}
\vskip -0.2in
\end{figure}

\rebuttal{
\subsection{MyoSuite Musculoskeletal Results}
\label{app:myosuite}

In this section, we evaluate \our in five musculoskeletal tasks from MyoSuite.

In these experiments, we followed \citet{hafner2023mastering,hansenTDMPC2ScalableRobust2023} and
scaled the rewards using $\mathrm{symlog}(\cdot)$,
\begin{equation}
  \mathrm{symlog}(x) = \mathrm{sign}(x)\mathrm{ln}(|x|+1).
\end{equation}
This compresses large and small rewards whilst preserving the input sign as it is a symmetric function.
Note that we simply transform the rewards with $\mathrm{symlog}$ and learn both the reward function and $Q\text{-functions}$
using these transformed rewards.
We use $N=1\text{-step}$ returns  in Hand Key Turn, Hand Obj Hold and Hand Pen Twirl and
we use $N=5\text{-step}$ returns in Hand Pose and Hand Reach.
In Hand Pose we also had to adjust the temperature from $0.5$ to $0.2$.
In future work, it would be interesting to investigate if using $\lambda\text{-returns}$ -- which uses a weighted-sum of $N\text{-step}$ returns -- can make
\our robust to the $N\text{-step}$ hyperparameter.
Further to this, it would be interesting to explore methods for dynamically tuning the MPPI (inverse) temperature $\tau_{\text{MPPI}}$.

In \cref{fig:myosuite} we show the training curves for the individual tasks.
\cref{fig:myosuite_rliable} then reports aggregate metrics at 1M environment steps over three random seeds in the five tasks.
On average, \our performs well, generally matching TD-MPC2 at 1M environment steps and outperforming the other baselines.

\begin{figure}[h]
\vskip 0.2in
\begin{center}
\centerline{\includegraphics[width=\textwidth]{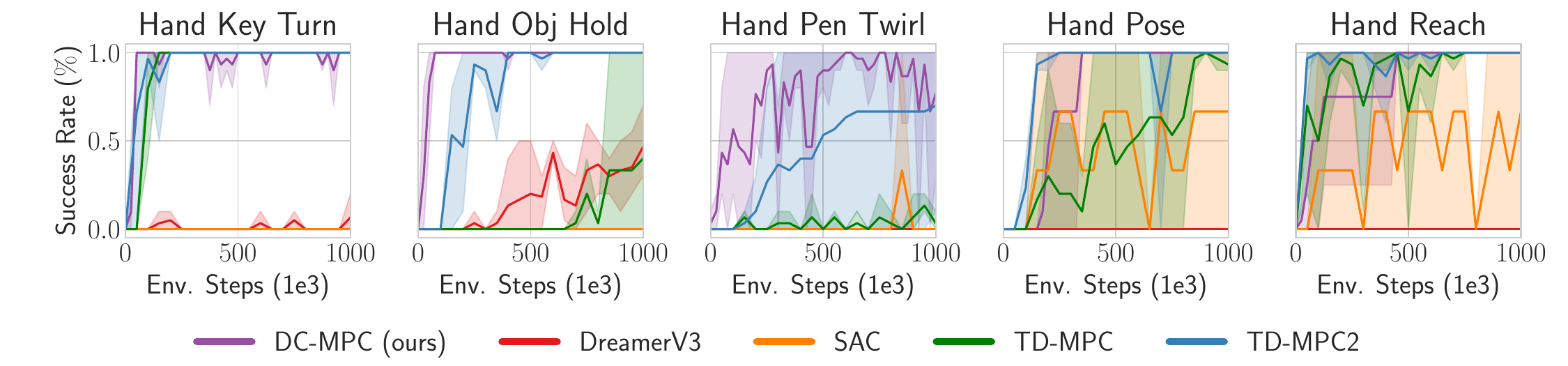}}
\caption{\textbf{MyoSuite training curves} We plot the mean (solid line) and the $95\%$ confidence intervals (shaded) across 3 seeds, where each seed averages over 10 evaluation episodes.}
\label{fig:myosuite}
\end{center}
\vskip -0.2in
\end{figure}
}

\vfill
\begin{center}
    --appendices continue on next page--
\end{center}

\clearpage

\rebuttal{
\subsection{Does \ourM Improve DreamerV3?}
\label{app:decoder-ablation}

In this section, we seek to evaluate what happens when we replace DreamerV3's one-hot discrete encoding
with the codebook encoding used in \our.
\cref{fig:dreamer-fsq-ablation} shows that in the easy Reacher Hard and Walker Walk environments, FSQ (blue) and
one-hot (orange) perform similarly.
However, in the difficult Dog Run and Humanoid Walk tasks, no discrete encoding can enable DreamerV3 to perform as well
as \our (purple).
We hypothesize that DreamerV3's poor performance in the Dog Run and Humanoid Walk tasks results from its decoder struggling
to reconstruct the observations.
\begin{figure}[h]
	\begin{center}
			\centerline{\includegraphics[width=0.9\textwidth]{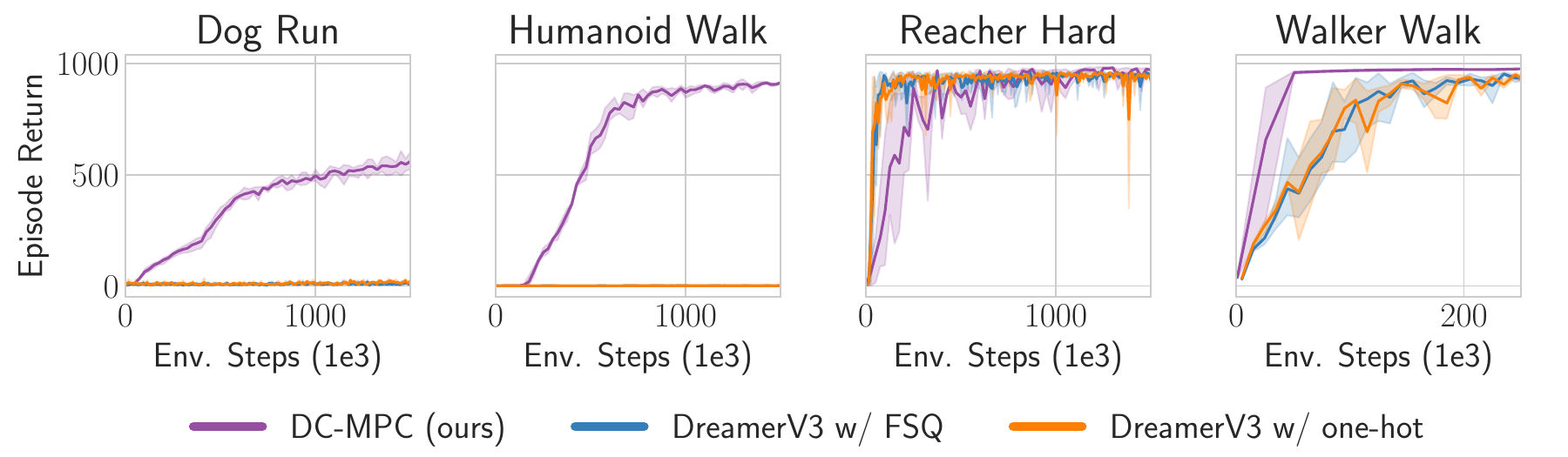}}
			\caption{\textbf{DreamerV3 with FSQ} Replacing DreamerV3's one-hot encoding (orange) with \our's codebook encoding (blue) does not improve performance. Moreover, DreamerV3 is not able to learn in the hard Dog Run and Humanoid Walk tasks and is significantly outperformed by \our (purple).}
			\label{fig:dreamer-fsq-ablation}
		\end{center}
        \vskip -0.2in
\end{figure}

Learning to minimize the observation reconstruction error has been widely applied in model-based RL \citep{sutton2018reinforcement, haRecurrentWorldModels2018, hafnerLearning2019}, and an observation decoder has been a component of many of the most successful RL algorithms to date \citep{hafner2023mastering}. However, recent work in representation learning for RL \citep{zhaoSimplifiedTemporalConsistency2023} and model-based RL \citep{hansenTemporalDifferenceLearning2022} has shown that incorporating a reconstruction term into the representation loss can hurt the performance, as learning to reconstruct the observations is inefficient due to the observations containing irrelevant details that are uncontrollable by the agent and do not affect the task.\looseness-1

To provide a thorough analysis of \our, we include results where we add a reconstruction term to our world model loss in \cref{eq:world-model-loss}:
\begin{align}
	\mathcal{L}_{\vo} = \mathbb{E}_{\vo_t\sim\mathcal{D}} [\| \hat{\vo}_t - \vo_t \|_2^2], \quad
\hat{\vo}_t = h_\kappa(\vc_t),
\end{align}
where $h_\kappa$ is a learned observation decoder that takes the latent code as the input and outputs the reconstructed observation.
The decoder $h_\kappa$ is a standard MLP. We perform reconstruction at each time step in the horizon.
The results in \cref{fig:decoder-ablation} show that in no environments does reconstruction aid learning, and in some tasks, such as the difficult Dog Run and Humanoid Walk tasks, including the reconstruction term has a significant detrimental effect on the performance, and can even prevent learning completely.
Our results support the observations of \citet{zhaoSimplifiedTemporalConsistency2023} and \citet{hansenTemporalDifferenceLearning2022} about the lack of need for a reconstruction loss in continuous control tasks.
However, it is worth noting that we weighted all loss terms equally whilst the results in
\citet{maHarmonyDreamTaskHarmonization2024} suggest that the observation reconstruction, temporal consistency, and reward
prediction loss terms need to be carefully balanced.\looseness-1
\begin{figure}[h]
	\begin{center}
			\centerline{\includegraphics[width=0.9\textwidth]{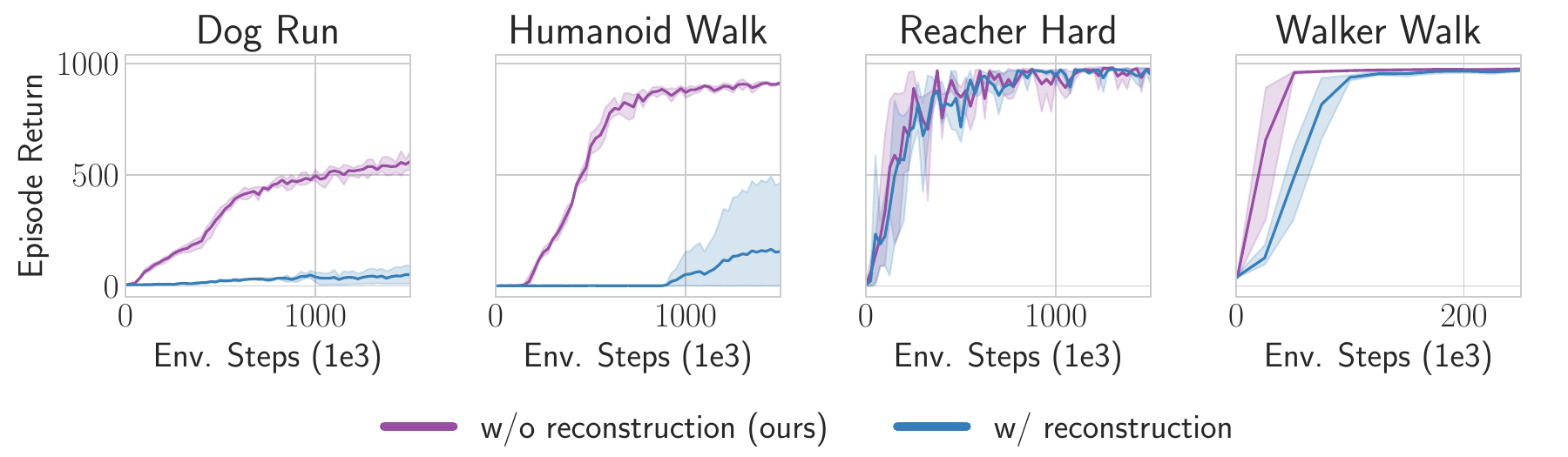}}
			\caption{\textbf{Reconstruction harms performance} Adding observation reconstruction to \our (blue) harms the performance of \our across a mixture of easy and hard DMControl tasks.
            }
			\label{fig:decoder-ablation}
		\end{center}
        \vskip -0.2in
\end{figure}
}

\rebuttal{
\subsection{Improving TD-MPC2 with \our}
\label{app:tdmpc2_with_dcwm}

In this section, we investigate using \our's latent space inside TD-MPC2.
Note that TD-MPC2's latent space is continuous and trained with MSE regression.
It also uses simplical normalization (SimNorm) to make its latent space bounded.
In these experiments, we removed SimNorm and replaced it with our discrete and stochastic latent space, and then trained using cross-entropy for the consistency loss.
In particular, we made the following changes to the TD-MPC2 codebase: {\em (i)} removed SimNorm, {\em (ii)} added FSQ to the encoder,
{\em (iii)} modified the dynamics to predict the logits instead of the next latent state,
{\em (iv)} modified the dynamics to use ST Gumbel-softmax sampling for multi-step predictions during training and
our weighted average approach during planning, and {\em (v)} changed the world model's loss coefficients  for consistency, value, and, reward, to all be $1$.

In \cref{fig:tdmpc2_with_dcwm_rliable},
we report aggregate metrics over $3$ random seeds in $10$ DMControl tasks and $10$ Meta-World tasks.
\cref{fig:tdmpc2_with_dcwm_rliable} (left) shows the IQM and optimality gap at 1M environment steps over the 20 tasks.
It shows that adding \our's discrete and stochastic latent space to TD-MPC2 offers some improvement.
\cref{fig:tdmpc2_with_dcwm_rliable} (right) shows the aggregate training curves (IQM over 10 tasks) for DMControl
and Meta-World, respectively.
The results show that using \ourM inside TD-MPC2 offers some benefits in the 10 DMControl tasks,
whilst in the 10 Meta-World tasks, the performance of all methods seems about equal.
This suggests that, in the context of continuous control, discrete and stochastic latent spaces are advantageous
for world models.
This is an interesting result which we believe motivates further research into discrete and stochastic latent spaces for world models.

}

\vfill
\begin{center}
    --appendices continue on next page--
\end{center}

\clearpage

\section{Implementation details}
\label{app:implementation}

\paragraph{Architecture}
We implemented \our with PyTorch \citep{paszkePyTorchImperativeStyle2019} and used the AdamW
optimizer \citep{KingmaB14} for training the models.
All components (encoder, dynamics, reward, actor and critic) are implemented as MLPs.
Following \citet{hansenTDMPC2ScalableRobust2023} we let all intermediate layers be linear layers followed by
LayerNorm \citep{baLayerNormalization2016}.
We use Mish activation functions throughout.
Below we summarize the \our architecture for our base model.

\begin{lstlisting}[basicstyle=\ttfamily\scriptsize]
DCMPC(
  (model): WorldModel(
    (_fsq): FSQ(levels=[5, 3])
    (_encoder): ModuleDict(
      (state): Sequential(
        (0): NormedLinear(in_features=obs_dim, out_features=256, act=Mish)
        (1): Linear(in_features=256, out_features=latent_dim*num_channels)
      )
    )
    (_trans): Sequential(
      (0): NormedLinear(in_features=(latent_dim*num_channels)+act_dim, out_features=512, act=Mish)
      (1): NormedLinear(in_features=512, out_features=512, act=Mish)
      (2): Linear(in_features=512, out_features=latent_dim*codebook_size)
    )
    (_reward): Sequential(
      (0): NormedLinear(in_features=(latent_dim*num_channels)+act_dim, out_features=512, act=Mish)
      (1): NormedLinear(in_features=512, out_features=512, act=Mish)
      (2): Linear(in_features=512, out_features=1)
    )
  )
  (_pi): Sequential(
    (0): NormedLinear(in_features=latent_dim*num_channels, out_features=512, act=Mish)
    (1): NormedLinear(in_features=512, out_features=512, act=Mish)
    (2): Linear(in_features=512, out_features=act_dim)
  )
  (_Qs): Vectorized ModuleList(
    (0-4): 5 x Sequential(
      (0): NormedLinear(in_features=(latent_dim*num_channels)+act_dim, out_features=512, act=Mish)
      (1): NormedLinear(in_features=512, out_features=512, act=Mish)
      (2): Linear(in_features=512, out_features=1)
    )
  )
  (_pi_tar): Sequential(
    (0): NormedLinear(in_features=latent_dim*num_channels, out_features=512, act=Mish)
    (1): NormedLinear(in_features=512, out_features=512, act=Mish)
    (2): Linear(in_features=512, out_features=act_dim)
  )
  (Qs_tar): Vectorized ModuleList(
    (0-4): 5 x Sequential(
      (0): NormedLinear(in_features=(latent_dim*num_channels)+act_dim, out_features=512, act=Mish)
      (1): NormedLinear(in_features=512, out_features=512, act=Mish)
      (2): Linear(in_features=512, out_features=1)
    )
  )
)
\end{lstlisting}
where \texttt{obs\_dim} is the dimensionality of the observation space,
\texttt{act\_dim} is the dimensionality of the action space,
\texttt{latent\_dim} is the number of the latent dimensions $d$ (default $512$),
\texttt{num\_channels} is the number of channels per latent dimension $b$ (default $2$),
and \texttt{codebook\_size} is the codebook size $|\mathcal{C}|$ (default $15$).

\paragraph{Statistical significance}
We used five seeds for \our and three seeds for TD-MPC2/DreamerV3/SAC/TD-MPC in the main figures, at least three seeds for all ablations, and plotted the
95 \% confidence intervals as the shaded area, which corresponds to approximately two standard errors of the mean.
However, in \cref{fig:latent-space-comparison,fig:tdmpc2_with_dcwm_rliable} we follow \citet{agarwalDeepReinforcementLearning2021} and plot the interquartile mean (IQM)
with the shaded area representing $95\%$ stratified bootstrap confidence intervals.

\paragraph{Hardware}
We used NVIDIA A100s and AMD Instinct MI250X GPUs to run our experiments. All our experiments have been run on a single GPU with a single-digit number of CPU workers.

\paragraph{Open-source code} For full details of the implementation, model architectures, and training, please check the code, which is available in the submitted supplementary material and available on github at
\url{https://github.com/aidanscannell/dcmpc}.

\vfill
\begin{center}
    --appendices continue on next page--
\end{center}

\clearpage

\paragraph{Hyperparameters}
\cref{tab:hyperparameters} lists all of the hyperparameters for training \our which were used for the main experiments
and the ablations.
\begin{table}[h]
\caption{\textbf{\our hyperparameters} We kept most hyperparameters fixed across tasks. However, we set task-specific exploration noise schedules and $N\text{-step}$ returns.}
\label{tab:hyperparameters}
\begin{center}
\resizebox{\textwidth}{!}{
\begin{footnotesize}
\begin{sc}
\begin{tabular}{lll}
\toprule
Hyperparameter & Value & Description \\
\midrule
\textbf{Training} & & \\
Action repeat & 2 (1 in MyoSuite) & \\
Max episode length $(T)$ & 500 in DMControl &  Action repeat makes this 1000 \\
 & 100 in Meta-World &  Action repeat makes this 200 \\
 & 100 in MyoSuite &  \\
Num. eval episodes & $10$ & \\
Random episodes $(N_{\text{random episodes}})$ & $10$ & Num. random episodes at start \\
\hline
\textbf{MPPI planning} & & \\
Planning horizon & $3$  & \\
Planning iterations ($J$) & $6$  & \\
Population size $(N_p)$ & $512$  & \\
Prior population size $(N_{\pi})$ & $24$  & Num. policy samples to warm start \\
Number of elites $(K)$ & $64$ & \\
Minimum std & $0.05$ & \\
Maximum std & $2$ & \\
(inverse) Temperature ($\tau_{\text{MPPI}}$) & $0.5$ & \\
\hline
\textbf{TD3} & & \\
Actor update freq. & $2$  & Update actor less than critic \\
Batch size & $512$ & \\
Buffer size & $10^{6}$ & \\
Discount factor $(\gamma)$ & $0.99$ & \\
Exploration noise & $\mathrm{Linear}(1.0,0.1,50)$ (easy) & DMControl \\
                  & $\mathrm{Linear}(1.0,0.1,150)$ (medium) & DMControl \\
                  & $\mathrm{Linear}(1.0,0.1,500)$ (hard) & DMControl \\
                  & $\mathrm{Linear}(1.0,0.1,250)$  & Meta-World \& MyoSuite \\
Learning rate & $3 \times 10^{-4}$ & \\
MLP dims & $[512, 512]$ & For actor/critic/dynamics/reward \\
Momentum coef. ($\tau$) & $0.005$ & \\
Num. $Q\text{-functions}$ ($N_{q}$) & $5$ & \\
Num. $Q\text{-functions}$ to sample & $2$ & \\
Noise clip $(c)$ & $0.3$ & \\
N-step TD & $1$ or $3$ in DMControl & \\
          & $3$ in Meta-World & \\
          & $1$ or $5$ in MyoSuite & \\
Policy noise & $0.2$ & \\
Update-to-data (UTD) ratio & $1$ & \\
\hline
\textbf{World model} &  & \\
Discount factor $(\gamma)$ & $0.9$ & \\
Encoder learning rate & $10^{-4}$ & \\
Encoder MLP dims & $[256]$ & \\
FSQ levels & $[5, 3]$ & Gives $|\mathcal{C}| = 5 \times 3 = 15 \approx 2^4$ \\
Horizon $(H)$ & $5$ & For world model training \\
Latent dimension ($d$) & $512$  & \\
                       & $1024$ (Humanoid/Dog) & \\
 \bottomrule
\end{tabular}
\end{sc}
\end{footnotesize}
}
\end{center}
\vskip -0.1in
\end{table}

\vfill
\begin{center}
    --appendices continue on next page--
\end{center}

\clearpage
\section{Baselines}
\label{app:baselines}
In this section, we provide further details of the baselines we compare against.

\begin{itemize}
  \item \textbf{DreamerV3 \citep{hafner2023mastering}} is a reinforcement learning algorithm that uses a world model to predict outcomes, a critic to judge their value, and an actor to choose actions to maximize value. 
  It uses symlog loss for training and operates on model states from imagination data. 
  The critic is a categorical distribution with exponentially spaced bins, and the actor is trained with entropy regularization and return normalization. 
  The world model is only used for training and there is no decision-time planning.
  In contrast, \our learns a deterministic encoder with a discrete latent space and stochastic dynamics in the world model. 
  We report the results of DreamerV3 from the TD-MPC2 official repository \footnote{\url{https://github.com/nicklashansen/tdmpc2/tree/main/results/dreamerv3}}.
  
  \item \textbf{Temporal Difference Model Predictive Control 2 (TD-MPC2, \citet{hansenTDMPC2ScalableRobust2023})} is a decoder-free model-based reinforcement learning algorithm with a focus on scalability and sample efficiency.
  It includes an encoder, latent transition dynamics, a reward predictor, a terminal value (critic), and a policy prior (actor). 
  In contrast to DreamerV3, it utilizes a deterministic encoder and transition dynamics implemented with MLPs, layer normalization \citep{baLayerNormalization2016} and Mish \citep{misra2019mish} activations.
  To avoid exploding gradients and representation collapse, the latent space is normalized with projection followed by a softmax operation. 
  All components except the policy prior are trained jointly based on predicting the latent embedding, reward prediction, and value prediction, while reward and value predictions are based on discrete regression in log-transformed space.
  Similarly, we use a deterministic encoder, but we train the transition dynamics with a cross-entropy loss function, which considers multi-modality and uncertainties, and we decouple representation learning from value learning. 
  We report the results from the TD-MPC2 official repository \footnote{\url{https://github.com/nicklashansen/tdmpc2/tree/main/results/tdmpc2}}.
    \rebuttal{
  \item \textbf{Temporal Difference Model Predictive Control (TD-MPC, \citet{hansenTemporalDifferenceLearning2022})} is the first version of TD-MPC2. It is also a decoder-free model-based RL algorithm consisting of an encoder, latent transition dynamics, reward predictor, terminal value (critic), and policy prior (actor).
    In contrast to TD-MPC2, it does not apply simplical normalization (SimNorm) to its latent state, it trains the reward and value prediction using the MSE loss instead of the cross-entropy loss, and it uses SAC as the underlying RL algorithm.
    We refer the reader to the TD-MPC paper for further details.
    We report the results from the TD-MPC2 official repository \footnote{\url{https://github.com/nicklashansen/tdmpc2/tree/main/results/tdmpc}}.\looseness-1

    \item \textbf{Soft Actor-Critic (SAC, \citet{haarnojaSoft2018} } is an off-policy model-free RL algorithm based on the maximum entropy RL framework. That is, it attempts to succeed at the task whilst acting as randomly as possible. It is worth highlighting that TD-MPC2 uses SAC as it's underlying model-free RL algorithm.
  We report the results from the TD-MPC2 official repository \footnote{\url{https://github.com/nicklashansen/tdmpc2/tree/main/results/sac}}.
    }

\end{itemize}

\vfill
\begin{center}
    --appendices continue on next page--
\end{center}

\clearpage
\section{Tasks}
\label{app:tasks}
We evaluate our method in 30 tasks from the DeepMind Control suite \citep{tassa2018deepmind}, 45 tasks from Meta-World \citep{yu2019meta} and 5 tasks from MyoSuite \citep{MyoSuite2022}.
\cref{tab:dmc_tasks,tab:mw_tasks,tab:myo_tasks} provide details of the environments we used, including the dimensionality of the observation
and action spaces.

\begin{table}[h]
\caption{\textbf{DMControl} We consider a total of 30 continuous control tasks from DMControl.}
\label{tab:dmc_tasks}
\begin{center}
\begin{sc}
\begin{footnotesize}

\begin{tabular}{lccc}
\toprule
\textbf{Task} & \textbf{Observation dim} & \textbf{Action dim}  & \textbf{Sparse?} \\
\midrule
Acrobot Swingup & 6 & 1 & N \\
Cartpole Balance & 5 & 1 & N \\
Carpole Balance Sparse & 5 & 1 & Y \\
Cartpole Swingup & 5 & 1 & N \\
Cartpole Swingup Sparse & 5 & 1 & Y \\
Cheetah Run & 17 & 6 & N \\
Cup Catch & 8 & 2 & Y \\
Cup Spin  & 8 & 2 & N \\
Dog Run & 223 & 38 & N \\
Dog Stand & 223 & 38 & N \\
Dog Trot & 223 & 38 & N \\
Dog Walk & 223 & 38 & N \\
Finger Spin & 9 & 2 & Y \\
Finger Turn Easy & 12 & 2 & Y \\
Finger Turn Hard & 12 & 2 & Y \\
Fish Swim & 24 & 5 & N \\
Hopper Hop & 15 & 4 & N \\
Hopper Stand & 15 & 4 & N \\
Humanoid Run & 67 & 24 & N \\
Humanoid Stand & 67 & 24 & N \\
Humanoid Walk & 67 & 24 & N \\
Pendulum Spin & 3 & 1 & N \\
Pendulum Swingup & 3 & 1 & N \\
Quadruped Run & 78 & 12 & N \\
Quadruped Walk & 78 & 12 & N \\
Reacher Easy & 6 & 2 & Y \\
Reacher Hard & 6 & 2 & Y \\
Walker Run & 24 & 6 & N \\
Walker Stand  & 24 & 6 & N \\
Walker Walk  & 24 & 6 & N \\
 \bottomrule
\end{tabular}
\end{footnotesize}
\end{sc}
\end{center}
\vskip -0.1in
\end{table}

\begin{table}[h]
\caption{\textbf{Meta-World} We consider a total of 45 continuous control tasks from Meta-World.
This benchmark is designed for multitask research so all tasks share similar embodiment, observation space, and action space.}
\label{tab:mw_tasks}
\begin{center}
\begin{sc}
\begin{footnotesize}

\begin{tabular}{lccc}
\toprule
\textbf{Task} & \textbf{Observation dim} & \textbf{Action dim}  & \textbf{Sparse?} \\
\midrule
Assembly & 39 & 4 & N \\
Basketball & 39 & 4 & N \\
Bin Picking & 39 & 4 & N \\
Box Close & 39 & 4 & N \\
Button Press & 39 & 4 & N \\
Button Press Topdown & 39 & 4 & N \\
Button Press Topdown Wall & 39 & 4 & N \\
Button Press Wall & 39 & 4 & N \\
Coffee Button & 39 & 4 & N \\
Coffee Push & 39 & 4 & N \\
Coffee Pull & 39 & 4 & N \\
Dial Turn & 39 & 4 & N \\
Disassemble & 39 & 4 & N \\
Door Close & 39 & 4 & N \\
Door Lock & 39 & 4 & N \\
Door Open & 39 & 4 & N \\
Door Unlock & 39 & 4 & N \\
Drawer Close & 39 & 4 & N \\
Faucet Close & 39 & 4 & N \\
Faucet Open & 39 & 4 & N \\
Hammer & 39 & 4 & N \\
Hand Insert & 39 & 4 & N \\
Handle Press & 39 & 4 & N \\
Handle Press Side & 39 & 4 & N \\
Handle Pull & 39 & 4 & N \\
Handle Pull Side & 39 & 4 & N \\
Lever Pull & 39 & 4 & N \\
Peg Insert Side & 39 & 4 & N \\
Peg Unplug Side & 39 & 4 & N \\
Pick Out Of Hole & 39 & 4 & N \\
Pick Place & 39 & 4 & N \\
Plate Slide & 39 & 4 & N \\
Plate Slide Back & 39 & 4 & N \\
Plate Slide Back Side & 39 & 4 & N \\
Plate Slide Side & 39 & 4 & N \\
Push & 39 & 4 & N \\
Push Wall & 39 & 4 & N \\
Reach Wall & 39 & 4 & N \\
Soccer & 39 & 4 & N \\
Stick Pull & 39 & 4 & N \\
Stick Push & 39 & 4 & N \\
Sweep & 39 & 4 & N \\
Sweep Into & 39 & 4 & N \\
Window Close & 39 & 4 & N \\
Window Open & 39 & 4 & N \\
 \bottomrule
\end{tabular}
\end{footnotesize}
\end{sc}
\end{center}
\vskip -0.1in
\end{table}

\begin{table}[h]
\caption{\textbf{MyoSuite} We consider a total of 5 continuous control tasks from MyoSuite.
  This benchmark is designed for high-dimensional muscoloskeletal motor control which involves complex object manipulation with a dexterous hand.}
\label{tab:myo_tasks}
\begin{center}
\begin{sc}
\begin{footnotesize}

\begin{tabular}{lccc}
\toprule
\textbf{Task} & \textbf{Observation dim} & \textbf{Action dim}  & \textbf{Sparse?} \\
\midrule
Key Turn & 93 & 39 & N \\
Object Hold & 91 & 39 & N \\
Pen Twirl & 83 & 39 & N \\
Pose & 108 & 39 & N \\
Reach & 115 & 39 & N \\
 \bottomrule
\end{tabular}
\end{footnotesize}
\end{sc}
\end{center}
\vskip -0.1in
\end{table}

\vfill
\begin{center}
    --appendices continue on next page--
\end{center}

\clearpage

\begin{figure}[t]
  \centering
  \includegraphics[width=\linewidth]{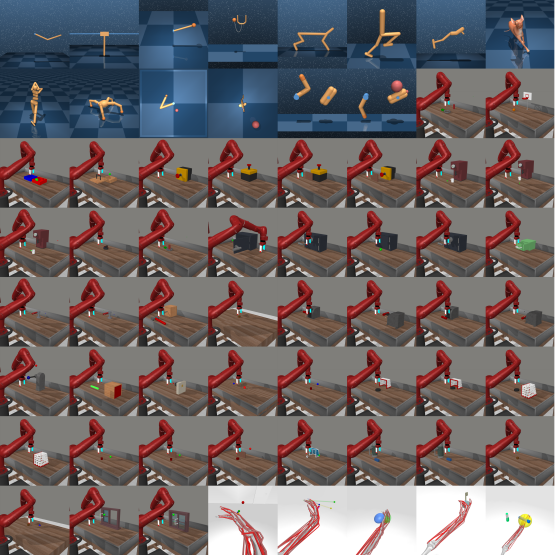}
  \caption{\textbf{Tasks visualizations} Visualization of the DMControl, Meta-World, and MyoSuite tasks used throughout the paper.}
  \label{fig:tasks}
\end{figure}

\end{document}

%% file: math_commands.tex
\usepackage{amsmath,amsfonts,bm}

\def\eqref#1{equation~\ref{#1}}

\def\1{\bm{1}}

\def\vmu{{\bm{\mu}}}

\def\va{{\bm{a}}}

\def\vc{{\bm{c}}}

\def\ve{{\bm{e}}}

\def\vl{{\bm{l}}}

\def\vo{{\bm{o}}}

\def\vq{{\bm{q}}}
\def\vr{{\bm{r}}}

\def\vv{{\bm{v}}}

\def\vx{{\bm{x}}}

\def\vz{{\bm{z}}}

\DeclareMathAlphabet{\mathsfit}{\encodingdefault}{\sfdefault}{m}{sl}
\SetMathAlphabet{\mathsfit}{bold}{\encodingdefault}{\sfdefault}{bx}{n}

\newcommand{\E}{\mathbb{E}}

\newcommand{\R}{\mathbb{R}}

\newcommand{\softmax}{\mathrm{softmax}}

\DeclareMathOperator*{\argmax}{arg\,max}

%% file: figs/hypercube_figure.tex
\begin{wrapfigure}{R}{.45\textwidth}
\vspace{-2em}
\centering
\begin{subfigure}[b]{.3\textwidth}

  \tdplotsetmaincoords{45}{135}

  \begin{tikzpicture}[tdplot_main_coords, scale=1,remember picture,scale=.75]

    \def\xsc{.5}
    \def\ysc{.5}
    \def\zsc{1}

    \tikzstyle{axis}=[->,thick,draw=blue!60!black]
    \tikzstyle{point}=[fill=white, draw=black, circle, inner sep=1.5pt]

    \draw[axis] (0,0,0) -- (3,0,0) node[anchor=north east]{$x$};
    \draw[axis] (0,0,0) -- (0,3,0) node[anchor=north west]{$y$};
    \draw[axis] (0,0,0) -- (0,0,3) node[anchor=south]{$z$};

    \foreach \x in {-2,-1,0,1,2}
        \foreach \y in {-2,-1,0,1,2}
            \foreach \z in {-1,-.333,.333,1}
                \node[point] (p-\x-\y-\z) at (\x*\xsc,\y*\ysc,\z*\zsc) {};

    \foreach \z in {-1,-.333,.333,1}
      \draw[fill=white,opacity=.6,draw=black!10,thin] (-1,-1,\z) -- (1,-1,\z) -- (1,1,\z) -- (-1,1,\z) -- cycle;

    \draw[draw=black,thin] (-1,-1,1) -- (1,-1,1) -- (1,1,1) -- (-1,1,1) -- cycle;
    \draw[draw=black,thin] (1,-1,-1) -- (1,-1,1) -- (1,1,1) -- (1,1,-1) -- cycle;
    \draw[draw=black,thin] (-1,1,-1) -- (-1,1,1) -- (1,1,1) -- (1,1,-1) -- cycle;

    \foreach \y in {-2,-1,0,1,2}
      \foreach \z in {-1,-.333,.333,1}
        \node[point] at (1,\y*\ysc,\z*\zsc) {};
    \foreach \x in {-2,-1,0,1,2}
      \foreach \z in {-1,-.333,.333,1}
        \node[point] at (\x*\xsc,1,\z*\zsc) {};
    \foreach \x in {-2,-1,0,1,2}
      \foreach \y in {-2,-1,0,1,2}
        \node[point] at (\x*\xsc,\y*\ysc,1) {};

    \draw[axis] (1,0,0) -- (3,0,0) node[anchor=north east]{$x$};
    \draw[axis] (0,1,0) -- (0,3,0) node[anchor=north west]{$y$};
    \draw[axis] (0,0,1) -- (0,0,3) node[anchor=south]{$z$};

  \end{tikzpicture}
\end{subfigure}
\begin{subfigure}[t]{.1\textwidth}
  \begin{tikzpicture}[remember picture]

    \tikzstyle{code}=[black!80,draw,inner sep=1pt,font=\scriptsize,minimum height=10pt,minimum width=10pt,inner sep=0,draw=black!80]

    \foreach \x/\y/\z [count=\i] in {-1/-1/1,-1/-.5/1,-1/1/.33}
      \foreach \u [count=\j] in {\x,\y,\z}
        \node[code] (c-\i-\j) at (\j*10pt,-\i*10pt) {\u};

    \node at (2*10pt,-4*10pt) {$\vdots$};
    \foreach \u [count=\j] in {1,2,3}
      \node[code] at (\j*10pt,-6*10pt) {};

    \node[align=center,anchor=north,font=\scriptsize] at (2*10pt,-7*10pt) {Fixed, implicit \\ codebook};

  \end{tikzpicture}
\end{subfigure}
%
\begin{tikzpicture}[remember picture,overlay]
  \tikzstyle{arr}=[draw=blue!80!white,thick, bend left=45,shorten >=2pt, shorten <=2pt]
  \draw[arr] (p--2--2-1) edge[->,bend left=20] (c-1-1);
  \draw[arr] (p--2--1-1) edge[->,bend left=20] (c-2-1);
  \draw[arr] ($(p--2-2-1)-(0,1em)$) edge[->,bend left=20] (c-3-1);
\end{tikzpicture}

\caption{\textbf{Illustration of Codebook ($\mathcal{C}$)} FSQ's codebook is a $b\text{-dimensional}$ hypercube (left). This figure illustrates a $b\text{=3-dimensional}$ codebook, where each axis of the $3\text{-dimensional}$ hypercube (left) corresponds to one dimension of the codebook (right). The $i^{\text{th}}$ dimension of the hypercube is discretized into $L_{i}$ values, \eg, the $x$ and $y\text{-axis}$ are discretized into $L_{0}=L_{1}=5$ and the $z\text{-axis}$ into $L_{3}=4$. Code symbols (here integers) are normalized to the range $[-1,1]$.}
\label{fig:fsq}
\vspace{-0.8em}
\end{wrapfigure}

%% file: arxiv.bbl
\begin{thebibliography}{68}
\providecommand{\natexlab}[1]{#1}
\providecommand{\url}[1]{\texttt{#1}}
\expandafter\ifx\csname urlstyle\endcsname\relax
  \providecommand{\doi}[1]{doi: #1}\else
  \providecommand{\doi}{doi: \begingroup \urlstyle{rm}\Url}\fi

\bibitem[Agarwal et~al.(2021)Agarwal, Schwarzer, Castro, Courville, and
  Bellemare]{agarwalDeepReinforcementLearning2021}
Rishabh Agarwal, Max Schwarzer, Pablo~Samuel Castro, Aaron~C Courville, and
  Marc Bellemare.
\newblock Deep {{Reinforcement Learning}} at the {{Edge}} of the {{Statistical
  Precipice}}.
\newblock In \emph{Advances in {{Neural Information Processing Systems}}},
  volume~34, pp.\  29304--29320. Curran Associates, Inc., 2021.

\bibitem[Allen \& Koomen(1983)Allen and Koomen]{allen1983planning}
James~F Allen and Johannes~A Koomen.
\newblock Planning using a temporal world model.
\newblock In \emph{Proceedings of the Eighth international joint conference on
  Artificial intelligence-Volume 2}, pp.\  741--747, 1983.

\bibitem[Alonso et~al.(2024)Alonso, Jelley, Micheli, Kanervisto, Storkey,
  Pearce, and Fleuret]{alonsoDiffusionWorldModeling2024}
Eloi Alonso, Adam Jelley, Vincent Micheli, Anssi Kanervisto, Amos Storkey, Tim
  Pearce, and Fran{\c c}ois Fleuret.
\newblock Diffusion for {{World Modeling}}: {{Visual Details Matter}} in
  {{Atari}}.
\newblock In \emph{The {{Thirty-eighth Annual Conference}} on {{Neural
  Information Processing Systems}}}, November 2024.

\bibitem[Ba et~al.(2016)Ba, Kiros, and Hinton]{baLayerNormalization2016}
Jimmy~Lei Ba, Jamie~Ryan Kiros, and Geoffrey~E. Hinton.
\newblock Layer {{Normalization}}.
\newblock \emph{arXiv preprint arXiv:1607.06450}, 2016.

\bibitem[Bar et~al.(2024)Bar, Zhou, Tran, Darrell, and
  LeCun]{barNavigationWorldModels2024}
Amir Bar, Gaoyue Zhou, Danny Tran, Trevor Darrell, and Yann LeCun.
\newblock Navigation {{World Models}}.
\newblock \emph{arXiv preprint arXiv:2412.03572}, 2024.

\bibitem[Basye et~al.(1992)Basye, Dean, Kirman, and Lejter]{basye1992decision}
Kenneth Basye, Thomas Dean, Jak Kirman, and Moises Lejter.
\newblock A decision-theoretic approach to planning, perception, and control.
\newblock \emph{IEEE Expert}, 7\penalty0 (4):\penalty0 58--65, 1992.

\bibitem[Bellman(1957)]{bellmanMarkovianDecisionProcess1957}
Richard Bellman.
\newblock A {{Markovian Decision Process}}.
\newblock \emph{Journal of Mathematics and Mechanics}, 6\penalty0 (5):\penalty0
  679--684, 1957.
\newblock ISSN 0095-9057.

\bibitem[Betts(1998)]{bettsSurvey1998}
John~T. Betts.
\newblock Survey of {{Numerical Methods}} for {{Trajectory Optimization}}.
\newblock \emph{Journal of Guidance, Control, and Dynamics}, 21\penalty0
  (2):\penalty0 193--207, March 1998.
\newblock \doi{10.2514/2.4231}.

\bibitem[Chang et~al.(2023)Chang, Zhang, Barber, Maschinot, Lezama, Jiang,
  Yang, Murphy, Freeman, Rubinstein, Li, and
  Krishnan]{changMuseTextToImageGeneration2023}
Huiwen Chang, Han Zhang, Jarred Barber, Aaron Maschinot, Jose Lezama, Lu~Jiang,
  Ming-Hsuan Yang, Kevin~Patrick Murphy, William~T. Freeman, Michael
  Rubinstein, Yuanzhen Li, and Dilip Krishnan.
\newblock Muse: {{Text-To-Image Generation}} via {{Masked Generative
  Transformers}}.
\newblock In \emph{Proceedings of the 40th {{International Conference}} on
  {{Machine Learning}}}, pp.\  4055--4075. PMLR, July 2023.

\bibitem[Chen et~al.(2021)Chen, Wang, Zhou, and
  Ross]{chenRandomizedEnsembledDouble2021}
Xinyue Chen, Che Wang, Zijian Zhou, and Keith Ross.
\newblock Randomized {{Ensembled Double Q-Learning}}: {{Learning Fast Without}}
  a {{Model}}.
\newblock In \emph{International {{Conference}} on {{Learning
  Representations}}}, 2021.

\bibitem[Chua et~al.(2018)Chua, Calandra, McAllister, and
  Levine]{chuaDeepReinforcementLearning2018}
Kurtland Chua, Roberto Calandra, Rowan McAllister, and Sergey Levine.
\newblock Deep {{Reinforcement Learning}} in a {{Handful}} of {{Trials}} using
  {{Probabilistic Dynamics Models}}.
\newblock In \emph{Advances in {{Neural Information Processing Systems}}},
  volume~31, 2018.

\bibitem[Daxberger et~al.(2021)Daxberger, Kristiadi, Immer, Eschenhagen, Bauer,
  and Hennig]{daxbergerLaplace2021}
Erik Daxberger, Agustinus Kristiadi, Alexander Immer, Runa Eschenhagen,
  Matthias Bauer, and Philipp Hennig.
\newblock Laplace {{Redux}} - {{Effortless Bayesian Deep Learning}}.
\newblock In \emph{Advances in {{Neural Information Processing Systems}}},
  volume~34, pp.\  20089--20103. Curran Associates, Inc., 2021.

\bibitem[Deng et~al.(2023)Deng, Park, and Ahn]{dengFacingWorldModel2023}
Fei Deng, Junyeong Park, and Sungjin Ahn.
\newblock Facing {{Off World Model Backbones}}: {{RNNs}}, {{Transformers}}, and
  {{S4}}.
\newblock In \emph{Advances in {{Neural Information Processing Systems}}},
  volume~36, pp.\  72904--72930, December 2023.

\bibitem[Esser et~al.(2021)Esser, Rombach, and
  Ommer]{esserTamingTransformersHighResolution2021}
Patrick Esser, Robin Rombach, and Bjorn Ommer.
\newblock Taming {{Transformers}} for {{High-Resolution Image Synthesis}}.
\newblock In \emph{Proceedings of the {{IEEE}}/{{CVF Conference}} on {{Computer
  Vision}} and {{Pattern Recognition}}}, pp.\  12873--12883, 2021.

\bibitem[Farebrother et~al.(2024)Farebrother, Orbay, Vuong, Ta{\"i}ga,
  Chebotar, Xiao, Irpan, Levine, Castro, Faust, Kumar, and
  Agarwal]{farebrotherStopRegressingTraining2024}
Jesse Farebrother, Jordi Orbay, Quan Vuong, Adrien~Ali Ta{\"i}ga, Yevgen
  Chebotar, Ted Xiao, Alex Irpan, Sergey Levine, Pablo~Samuel Castro,
  Aleksandra Faust, Aviral Kumar, and Rishabh Agarwal.
\newblock Stop {{Regressing}}: {{Training Value Functions}} via
  {{Classification}} for {{Scalable Deep RL}}, March 2024.

\bibitem[Fujimoto et~al.(2018)Fujimoto, Hoof, and
  Meger]{fujimotoAddressingFunctionApproximation2018}
Scott Fujimoto, Herke Hoof, and David Meger.
\newblock Addressing {{Function Approximation Error}} in {{Actor-Critic
  Methods}}.
\newblock In \emph{Proceedings of the 35th {{International Conference}} on
  {{Machine Learning}}}, pp.\  1587--1596. PMLR, July 2018.

\bibitem[Georgiev et~al.(2024)Georgiev, Giridhar, Hansen, and
  Garg]{georgiev2024pwm}
Ignat Georgiev, Varun Giridhar, Nicklas Hansen, and Animesh Garg.
\newblock {PWM}: {P}olicy {L}earning with {L}arge {W}orld {M}odels.
\newblock \emph{arXiv preprint 2407.02466}, 2024.

\bibitem[Ghugare et~al.(2022)Ghugare, Bharadhwaj, Eysenbach, Levine, and
  Salakhutdinov]{ghugareSimplifyingModelbasedRL2022a}
Raj Ghugare, Homanga Bharadhwaj, Benjamin Eysenbach, Sergey Levine, and Russ
  Salakhutdinov.
\newblock Simplifying {{Model-based RL}}: {{Learning Representations}},
  {{Latent-space Models}}, and {{Policies}} with {{One Objective}}.
\newblock In \emph{The {{Eleventh International Conference}} on {{Learning
  Representations}}}, September 2022.

\bibitem[Ha \& Schmidhuber(2018)Ha and Schmidhuber]{haRecurrentWorldModels2018}
David Ha and J{\"u}rgen Schmidhuber.
\newblock Recurrent {{World Models Facilitate Policy Evolution}}.
\newblock In \emph{Advances in {{Neural Information Processing Systems}}},
  volume~31. Curran Associates, Inc., 2018.

\bibitem[Haarnoja et~al.(2018)Haarnoja, Zhou, Abbeel, and
  Levine]{haarnojaSoft2018}
Tuomas Haarnoja, Aurick Zhou, Pieter Abbeel, and Sergey Levine.
\newblock Soft {{Actor-Critic}}: {{Off-Policy Maximum Entropy Deep
  Reinforcement Learning}} with a {{Stochastic Actor}}.
\newblock In \emph{International {{Conference}} on {{Machine Learning}}}, pp.\
  1861--1870. PMLR, July 2018.

\bibitem[Hafner et~al.(2019{\natexlab{a}})Hafner, Lillicrap, Ba, and
  Norouzi]{hafner2019dream}
Danijar Hafner, Timothy Lillicrap, Jimmy Ba, and Mohammad Norouzi.
\newblock Dream to control: Learning behaviors by latent imagination.
\newblock \emph{arXiv preprint arXiv:1912.01603}, 2019{\natexlab{a}}.

\bibitem[Hafner et~al.(2019{\natexlab{b}})Hafner, Lillicrap, Fischer, Villegas,
  Ha, Lee, and Davidson]{hafnerLearning2019}
Danijar Hafner, Timothy Lillicrap, Ian Fischer, Ruben Villegas, David Ha,
  Honglak Lee, and James Davidson.
\newblock Learning {{Latent Dynamics}} for {{Planning}} from {{Pixels}}.
\newblock In \emph{International {{Conference}} on {{Machine Learning}}}, pp.\
  2555--2565. PMLR, May 2019{\natexlab{b}}.

\bibitem[Hafner et~al.(2022)Hafner, Lillicrap, Norouzi, and
  Ba]{hafnerMasteringAtariDiscrete2022}
Danijar Hafner, Timothy~P. Lillicrap, Mohammad Norouzi, and Jimmy Ba.
\newblock Mastering {{Atari}} with {{Discrete World Models}}.
\newblock In \emph{International {{Conference}} on {{Learning
  Representations}}}, February 2022.

\bibitem[Hafner et~al.(2023)Hafner, Pasukonis, Ba, and
  Lillicrap]{hafner2023mastering}
Danijar Hafner, Jurgis Pasukonis, Jimmy Ba, and Timothy Lillicrap.
\newblock Mastering diverse domains through world models.
\newblock \emph{arXiv preprint arXiv:2301.04104}, 2023.

\bibitem[Hansen et~al.(2023)Hansen, Su, and
  Wang]{hansenTDMPC2ScalableRobust2023}
Nicklas Hansen, Hao Su, and Xiaolong Wang.
\newblock {{TD-MPC2}}: {{Scalable}}, {{Robust World Models}} for {{Continuous
  Control}}.
\newblock In \emph{The {{Twelfth International Conference}} on {{Learning
  Representations}}}, October 2023.

\bibitem[Hansen et~al.(2022)Hansen, Su, and
  Wang]{hansenTemporalDifferenceLearning2022}
Nicklas~A. Hansen, Hao Su, and Xiaolong Wang.
\newblock Temporal {{Difference Learning}} for {{Model Predictive Control}}.
\newblock In \emph{Proceedings of the 39th {{International Conference}} on
  {{Machine Learning}}}, pp.\  8387--8406. PMLR, June 2022.

\bibitem[Henighan et~al.(2020)Henighan, Kaplan, Katz, Chen, Hesse, Jackson,
  Jun, Brown, Dhariwal, Gray, Hallacy, Mann, Radford, Ramesh, Ryder, Ziegler,
  Schulman, Amodei, and McCandlish]{henighanScalingLawsAutoregressive2020}
Tom Henighan, Jared Kaplan, Mor Katz, Mark Chen, Christopher Hesse, Jacob
  Jackson, Heewoo Jun, Tom~B. Brown, Prafulla Dhariwal, Scott Gray, Chris
  Hallacy, Benjamin Mann, Alec Radford, Aditya Ramesh, Nick Ryder, Daniel~M.
  Ziegler, John Schulman, Dario Amodei, and Sam McCandlish.
\newblock Scaling {{Laws}} for {{Autoregressive Generative Modeling}}, November
  2020.

\bibitem[Ho et~al.(2020)Ho, Jain, and
  Abbeel]{hoDenoisingDiffusionProbabilistic2020}
Jonathan Ho, Ajay Jain, and Pieter Abbeel.
\newblock Denoising {{Diffusion Probabilistic Models}}.
\newblock In \emph{Advances in {{Neural Information Processing Systems}}},
  volume~33, pp.\  6840--6851. Curran Associates, Inc., 2020.

\bibitem[Hoffmann et~al.(2022)Hoffmann, Borgeaud, Mensch, Buchatskaya, Cai,
  Rutherford, Casas, Hendricks, Welbl, Clark, Hennigan, Noland, Millican,
  van~den Driessche, Damoc, Guy, Osindero, Simonyan, Elsen, Rae, Vinyals, and
  Sifre]{hoffmannTrainingComputeOptimalLarge2022}
Jordan Hoffmann, Sebastian Borgeaud, Arthur Mensch, Elena Buchatskaya, Trevor
  Cai, Eliza Rutherford, Diego de~Las Casas, Lisa~Anne Hendricks, Johannes
  Welbl, Aidan Clark, Tom Hennigan, Eric Noland, Katie Millican, George van~den
  Driessche, Bogdan Damoc, Aurelia Guy, Simon Osindero, Karen Simonyan, Erich
  Elsen, Jack~W. Rae, Oriol Vinyals, and Laurent Sifre.
\newblock Training {{Compute-Optimal Large Language Models}}, March 2022.

\bibitem[Hsu et~al.(2023)Hsu, Dorrell, Whittington, Wu, and
  Finn]{hsuDisentanglementLatentQuantization2023}
Kyle Hsu, William Dorrell, James Whittington, Jiajun Wu, and Chelsea Finn.
\newblock Disentanglement via {{Latent Quantization}}.
\newblock \emph{Advances in Neural Information Processing Systems},
  36:\penalty0 45463--45488, December 2023.

\bibitem[Igl et~al.(2018)Igl, Zintgraf, Le, Wood, and Whiteson]{igl2018deep}
Maximilian Igl, Luisa Zintgraf, Tuan~Anh Le, Frank Wood, and Shimon Whiteson.
\newblock Deep variational reinforcement learning for pomdps.
\newblock In \emph{International Conference on Machine Learning}, pp.\
  2117--2126. PMLR, 2018.

\bibitem[Jang et~al.(2017)Jang, Gu, and Poole]{jang2017categorical}
Eric Jang, Shixiang Gu, and Ben Poole.
\newblock Categorical reparameterization with gumbel-softmax.
\newblock In \emph{International Conference on Learning Representations}, 2017.

\bibitem[Kaplan et~al.(2020)Kaplan, McCandlish, Henighan, Brown, Chess, Child,
  Gray, Radford, Wu, and Amodei]{kaplanScalingLawsNeural2020}
Jared Kaplan, Sam McCandlish, Tom Henighan, Tom~B. Brown, Benjamin Chess, Rewon
  Child, Scott Gray, Alec Radford, Jeffrey Wu, and Dario Amodei.
\newblock Scaling {{Laws}} for {{Neural Language Models}}.
\newblock \emph{arXiv preprint arXiv:2001.08361}, 2020.

\bibitem[Kingma \& Ba(2015)Kingma and Ba]{KingmaB14}
Diederik~P. Kingma and Jimmy Ba.
\newblock Adam: A method for stochastic optimization.
\newblock In \emph{International Conference on Learning Representations}, 2015.

\bibitem[Kingma \& Welling(2014)Kingma and Welling]{kingmaAutoEncoding2014}
Diederik~P. Kingma and M.~Welling.
\newblock Auto-{{Encoding Variational Bayes}}.
\newblock In \emph{International Conference on Learning Representations}, 2014.

\bibitem[Kostrikov et~al.(2020)Kostrikov, Yarats, and
  Fergus]{kostrikov2020image}
Ilya Kostrikov, Denis Yarats, and Rob Fergus.
\newblock Image augmentation is all you need: Regularizing deep reinforcement
  learning from pixels.
\newblock \emph{arXiv preprint arXiv:2004.13649}, 2020.

\bibitem[LeCun()]{lecunPathAutonomousMachine}
Yann LeCun.
\newblock A {{Path Towards Autonomous Machine Intelligence Version}} 0.9.2,
  2022-06-27.

\bibitem[Lutter et~al.(2021)Lutter, Hasenclever, Byravan, Dulac-Arnold,
  Trochim, Heess, Merel, and Tassa]{lutter2021learning}
Michael Lutter, Leonard Hasenclever, Arunkumar Byravan, Gabriel Dulac-Arnold,
  Piotr Trochim, Nicolas Heess, Josh Merel, and Yuval Tassa.
\newblock Learning dynamics models for model predictive agents.
\newblock \emph{arXiv preprint arXiv:2109.14311}, 2021.

\bibitem[Ma et~al.(2024)Ma, Wu, Feng, Xiao, Li, Hao, Wang, and
  Long]{maHarmonyDreamTaskHarmonization2024}
Haoyu Ma, Jialong Wu, Ningya Feng, Chenjun Xiao, Dong Li, Jianye Hao, Jianmin
  Wang, and Mingsheng Long.
\newblock {{HarmonyDream}}: {{Task Harmonization Inside World Models}}.
\newblock In \emph{Proceedings of the 41st {{International Conference}} on
  {{Machine Learning}}}, pp.\  33983--34007. PMLR, July 2024.

\bibitem[Maddison et~al.(2017)Maddison, Mnih, and Teh]{maddison2017the}
Chris~J. Maddison, Andriy Mnih, and Yee~Whye Teh.
\newblock The concrete distribution: A continuous relaxation of discrete random
  variables.
\newblock In \emph{International Conference on Learning Representations}, 2017.

\bibitem[Mentzer et~al.(2024)Mentzer, Minnen, Agustsson, and
  Tschannen]{mentzerFiniteScalarQuantization2023}
Fabian Mentzer, David Minnen, Eirikur Agustsson, and Michael Tschannen.
\newblock Finite {{Scalar Quantization}}: {{VQ-VAE Made Simple}}.
\newblock In \emph{International Conference on Learning Representations}, 2024.

\bibitem[Micheli et~al.(2022)Micheli, Alonso, and
  Fleuret]{micheliTransformersAreSampleEfficient2022}
Vincent Micheli, Eloi Alonso, and Fran{\c c}ois Fleuret.
\newblock Transformers are {{Sample-Efficient World Models}}.
\newblock In \emph{The {{Eleventh International Conference}} on {{Learning
  Representations}}}, September 2022.

\bibitem[Misra(2019)]{misra2019mish}
Diganta Misra.
\newblock Mish: A self regularized non-monotonic activation function.
\newblock \emph{arXiv preprint arXiv:1908.08681}, 2019.

\bibitem[NVIDIA et~al.(2025)NVIDIA, :, Agarwal, Ali, Bala, Balaji, Barker, Cai,
  Chattopadhyay, Chen, Cui, Ding, Dworakowski, Fan, Fenzi, Ferroni, Fidler,
  Fox, Ge, Ge, Gu, Gururani, He, Huang, Huffman, Jannaty, Jin, Kim, Klár, Lam,
  Lan, Leal-Taixe, Li, Li, Lin, Lin, Ling, Liu, Liu, Luo, Ma, Mao, Mo,
  Mousavian, Nah, Niverty, Page, Paschalidou, Patel, Pavao, Ramezanali, Reda,
  Ren, Sabavat, Schmerling, Shi, Stefaniak, Tang, Tchapmi, Tredak, Tseng,
  Varghese, Wang, Wang, Wang, Wang, Wei, Wei, Wu, Xu, Yang, Yen-Chen, Zeng,
  Zeng, Zhang, Zhang, Zhang, Zhao, and
  Zolkowski]{nvidia2025cosmosworldfoundationmodel}
NVIDIA, :, Niket Agarwal, Arslan Ali, Maciej Bala, Yogesh Balaji, Erik Barker,
  Tiffany Cai, Prithvijit Chattopadhyay, Yongxin Chen, Yin Cui, Yifan Ding,
  Daniel Dworakowski, Jiaojiao Fan, Michele Fenzi, Francesco Ferroni, Sanja
  Fidler, Dieter Fox, Songwei Ge, Yunhao Ge, Jinwei Gu, Siddharth Gururani,
  Ethan He, Jiahui Huang, Jacob Huffman, Pooya Jannaty, Jingyi Jin, Seung~Wook
  Kim, Gergely Klár, Grace Lam, Shiyi Lan, Laura Leal-Taixe, Anqi Li, Zhaoshuo
  Li, Chen-Hsuan Lin, Tsung-Yi Lin, Huan Ling, Ming-Yu Liu, Xian Liu, Alice
  Luo, Qianli Ma, Hanzi Mao, Kaichun Mo, Arsalan Mousavian, Seungjun Nah,
  Sriharsha Niverty, David Page, Despoina Paschalidou, Zeeshan Patel, Lindsey
  Pavao, Morteza Ramezanali, Fitsum Reda, Xiaowei Ren, Vasanth Rao~Naik
  Sabavat, Ed~Schmerling, Stella Shi, Bartosz Stefaniak, Shitao Tang, Lyne
  Tchapmi, Przemek Tredak, Wei-Cheng Tseng, Jibin Varghese, Hao Wang, Haoxiang
  Wang, Heng Wang, Ting-Chun Wang, Fangyin Wei, Xinyue Wei, Jay~Zhangjie Wu,
  Jiashu Xu, Wei Yang, Lin Yen-Chen, Xiaohui Zeng, Yu~Zeng, Jing Zhang,
  Qinsheng Zhang, Yuxuan Zhang, Qingqing Zhao, and Artur Zolkowski.
\newblock Cosmos world foundation model platform for physical ai.
\newblock \emph{arXiv preprint arXiv:2501.03575}, 2025.

\bibitem[Paszke et~al.(2019)Paszke, Gross, Massa, Lerer, Bradbury, Chanan,
  Killeen, Lin, Gimelshein, Antiga, Desmaison, Kopf, Yang, DeVito, Raison,
  Tejani, Chilamkurthy, Steiner, Fang, Bai, and
  Chintala]{paszkePyTorchImperativeStyle2019}
Adam Paszke, Sam Gross, Francisco Massa, Adam Lerer, James Bradbury, Gregory
  Chanan, Trevor Killeen, Zeming Lin, Natalia Gimelshein, Luca Antiga, Alban
  Desmaison, Andreas Kopf, Edward Yang, Zachary DeVito, Martin Raison, Alykhan
  Tejani, Sasank Chilamkurthy, Benoit Steiner, Lu~Fang, Junjie Bai, and Soumith
  Chintala.
\newblock {{PyTorch}}: {{An Imperative Style}}, {{High-Performance Deep
  Learning Library}}.
\newblock In \emph{Advances in {{Neural Information Processing Systems}}},
  volume~32. Curran Associates, Inc., 2019.

\bibitem[Ramesh et~al.(2021)Ramesh, Pavlov, Goh, Gray, Voss, Radford, Chen, and
  Sutskever]{rameshZeroShotTexttoImageGeneration2021}
Aditya Ramesh, Mikhail Pavlov, Gabriel Goh, Scott Gray, Chelsea Voss, Alec
  Radford, Mark Chen, and Ilya Sutskever.
\newblock Zero-{{Shot Text-to-Image Generation}}.
\newblock In \emph{Proceedings of the 38th {{International Conference}} on
  {{Machine Learning}}}, pp.\  8821--8831. PMLR, July 2021.

\bibitem[Reed et~al.(2022)Reed, Zolna, Parisotto, Colmenarejo, Novikov,
  Barth-Maron, Gimenez, Sulsky, Kay, Springenberg, et~al.]{reed2022generalist}
Scott Reed, Konrad Zolna, Emilio Parisotto, Sergio~Gomez Colmenarejo, Alexander
  Novikov, Gabriel Barth-Maron, Mai Gimenez, Yury Sulsky, Jackie Kay,
  Jost~Tobias Springenberg, et~al.
\newblock {A Generalist Agent}.
\newblock \emph{Transactions on Machine Learning Research (TMLR)}, 2022.

\bibitem[Robine et~al.(2022)Robine, H{\"o}ftmann, Uelwer, and
  Harmeling]{robineTransformerbasedWorldModels2022}
Jan Robine, Marc H{\"o}ftmann, Tobias Uelwer, and Stefan Harmeling.
\newblock Transformer-based {{World Models Are Happy With}} 100k
  {{Interactions}}.
\newblock In \emph{The {{Eleventh International Conference}} on {{Learning
  Representations}}}, September 2022.

\bibitem[Rubinstein(1997)]{rubinstein1997optimization}
Reuven~Y Rubinstein.
\newblock Optimization of computer simulation models with rare events.
\newblock \emph{European Journal of Operational Research}, 99\penalty0
  (1):\penalty0 89--112, 1997.

\bibitem[Scannell et~al.(2021)Scannell, Ek, and
  Richards]{scannellTrajectory2021}
Aidan Scannell, Carl~Henrik Ek, and Arthur Richards.
\newblock Trajectory {{Optimisation}} in {{Learned Multimodal Dynamical Systems
  Via Latent-ODE Collocation}}.
\newblock In \emph{Proceedings of the {{IEEE International Conference}} on
  {{Robotics}} and {{Automation}}}. IEEE, 2021.

\bibitem[Scannell et~al.(2024{\natexlab{a}})Scannell, Kujanp{\"a}{\"a}, Zhao,
  Nakhaeinezhadfard, Solin, and
  Pajarinen]{scannellQuantizedRepresentationsPrevent2024}
Aidan Scannell, Kalle Kujanp{\"a}{\"a}, Yi~Zhao, Mohammadreza
  Nakhaeinezhadfard, Arno Solin, and Joni Pajarinen.
\newblock Quantized {{Representations Prevent Dimensional Collapse}} in
  {{Self-predictive RL}}.
\newblock In \emph{{{ICML}} 2024 {{Workshop}}: {{Aligning Reinforcement
  Learning Experimentalists}} and {{Theorists}}}, July 2024{\natexlab{a}}.

\bibitem[Scannell et~al.(2024{\natexlab{b}})Scannell, Kujanpää, Zhao,
  Nakhaei, Solin, and Pajarinen]{scannell2024iqrl}
Aidan Scannell, Kalle Kujanpää, Yi~Zhao, Mohammadreza Nakhaei, Arno Solin,
  and Joni Pajarinen.
\newblock {iQRL} - {I}mplicitly {Q}uantized {R}epresentations for
  {S}ample-efficient {R}einforcement {L}earning.
\newblock \emph{arXiv preprint arXiv:2406.02696}, 2024{\natexlab{b}}.

\bibitem[Scannell et~al.(2024{\natexlab{c}})Scannell, Mereu, Chang, Tamir,
  Pajarinen, and Solin]{scannell2024functionspace}
Aidan Scannell, Riccardo Mereu, Paul~Edmund Chang, Ella Tamir, Joni Pajarinen,
  and Arno Solin.
\newblock Function-space parameterization of neural networks for sequential
  learning.
\newblock In \emph{The Twelfth International Conference on Learning
  Representations}, 2024{\natexlab{c}}.

\bibitem[Schrittwieser et~al.(2020)Schrittwieser, Antonoglou, Hubert, Simonyan,
  Sifre, Schmitt, Guez, Lockhart, Hassabis, Graepel, Lillicrap, and
  Silver]{schrittwieserMastering2020}
Julian Schrittwieser, Ioannis Antonoglou, Thomas Hubert, Karen Simonyan,
  Laurent Sifre, Simon Schmitt, Arthur Guez, Edward Lockhart, Demis Hassabis,
  Thore Graepel, Timothy Lillicrap, and David Silver.
\newblock Mastering {{Atari}}, {{Go}}, {{Chess}} and {{Shogi}} by {{Planning}}
  with a {{Learned Model}}.
\newblock \emph{Nature}, 588\penalty0 (7839):\penalty0 604--609, December 2020.
\newblock ISSN 0028-0836, 1476-4687.
\newblock \doi{10.1038/s41586-020-03051-4}.

\bibitem[Schwarzer et~al.(2020)Schwarzer, Anand, Goel, Hjelm, Courville, and
  Bachman]{schwarzerDataEfficientReinforcementLearning2020}
Max Schwarzer, Ankesh Anand, Rishab Goel, R.~Devon Hjelm, Aaron Courville, and
  Philip Bachman.
\newblock Data-{{Efficient Reinforcement Learning}} with {{Self-Predictive
  Representations}}.
\newblock In \emph{International {{Conference}} on {{Learning
  Representations}}}, October 2020.

\bibitem[Sutton \& Barto(2018)Sutton and Barto]{sutton2018reinforcement}
R.S. Sutton and A.G. Barto.
\newblock \emph{Reinforcement Learning, Second Edition: {{An}} Introduction}.
\newblock Adaptive Computation and Machine Learning Series. MIT Press, 2018.
\newblock ISBN 978-0-262-35270-3.

\bibitem[Tassa et~al.(2018)Tassa, Doron, Muldal, Erez, Li, Casas, Budden,
  Abdolmaleki, Merel, Lefrancq, et~al.]{tassa2018deepmind}
Yuval Tassa, Yotam Doron, Alistair Muldal, Tom Erez, Yazhe Li, Diego de~Las
  Casas, David Budden, Abbas Abdolmaleki, Josh Merel, Andrew Lefrancq, et~al.
\newblock Deepmind control suite.
\newblock \emph{arXiv preprint arXiv:1801.00690}, 2018.

\bibitem[{van den Oord} et~al.(2017){van den Oord}, Vinyals, and
  {kavukcuoglu}]{vandenoordNeuralDiscreteRepresentation2017}
Aaron {van den Oord}, Oriol Vinyals, and koray {kavukcuoglu}.
\newblock Neural {{Discrete Representation Learning}}.
\newblock In \emph{Advances in {{Neural Information Processing Systems}}},
  volume~30. Curran Associates, Inc., 2017.

\bibitem[Vaswani et~al.(2017)Vaswani, Shazeer, Parmar, Uszkoreit, Jones, Gomez,
  ukasz Kaiser, and Polosukhin]{vaswaniAttentionAllYou2017}
Ashish Vaswani, Noam Shazeer, Niki Parmar, Jakob Uszkoreit, Llion Jones,
  Aidan~N Gomez, {\L}~ukasz Kaiser, and Illia Polosukhin.
\newblock Attention is {{All}} you {{Need}}.
\newblock In \emph{Advances in {{Neural Information Processing Systems}}},
  volume~30. Curran Associates, Inc., 2017.

\bibitem[Vittorio et~al.(2022)Vittorio, Huawei, Guillaume, Massimo, and
  Vikash]{MyoSuite2022}
Caggiano Vittorio, Wang Huawei, Durandau Guillaume, Sartori Massimo, and Kumar
  Vikash.
\newblock Myosuite -- a contact-rich simulation suite for musculoskeletal motor
  control.
\newblock \emph{arXiv preprint arXiv:2205.13600}, 2022.

\bibitem[Wang et~al.(2022)Wang, Du, Torralba, Isola, Zhang, and
  Tian]{wangDenoisedMDPsLearning2022}
Tongzhou Wang, Simon Du, Antonio Torralba, Phillip Isola, Amy Zhang, and
  Yuandong Tian.
\newblock Denoised {{MDPs}}: {{Learning World Models Better Than}} the {{World
  Itself}}.
\newblock In \emph{Proceedings of the 39th {{International Conference}} on
  {{Machine Learning}}}, pp.\  22591--22612. PMLR, June 2022.

\bibitem[Williams et~al.(2015)Williams, Aldrich, and
  Theodorou]{williams2015model}
Grady Williams, Andrew Aldrich, and Evangelos Theodorou.
\newblock Model predictive path integral control using covariance variable
  importance sampling.
\newblock \emph{arXiv preprint arXiv:1509.01149}, 2015.

\bibitem[Yarats et~al.(2021{\natexlab{a}})Yarats, Fergus, Lazaric, and
  Pinto]{yarats2021mastering}
Denis Yarats, Rob Fergus, Alessandro Lazaric, and Lerrel Pinto.
\newblock Mastering visual continuous control: Improved data-augmented
  reinforcement learning.
\newblock \emph{arXiv preprint arXiv:2107.09645}, 2021{\natexlab{a}}.

\bibitem[Yarats et~al.(2021{\natexlab{b}})Yarats, Fergus, Lazaric, and
  Pinto]{yaratsMasteringVisualContinuous2021}
Denis Yarats, Rob Fergus, Alessandro Lazaric, and Lerrel Pinto.
\newblock Mastering {{Visual Continuous Control}}: {{Improved Data-Augmented
  Reinforcement Learning}}.
\newblock In \emph{International {{Conference}} on {{Learning
  Representations}}}, October 2021{\natexlab{b}}.

\bibitem[Yu et~al.(2019)Yu, Quillen, He, Julian, Hausman, Finn, and
  Levine]{yu2019meta}
Tianhe Yu, Deirdre Quillen, Zhanpeng He, Ryan Julian, Karol Hausman, Chelsea
  Finn, and Sergey Levine.
\newblock Meta-world: A benchmark and evaluation for multi-task and meta
  reinforcement learning.
\newblock In \emph{Conference on Robot Learning (CoRL)}, 2019.

\bibitem[Zhang et~al.(2023)Zhang, Wang, Sun, Yuan, and
  Huang]{zhangSTORMEfficientStochastic2023}
Weipu Zhang, Gang Wang, Jian Sun, Yetian Yuan, and Gao Huang.
\newblock {{STORM}}: {{Efficient Stochastic Transformer}} based {{World
  Models}} for {{Reinforcement Learning}}.
\newblock In \emph{Advances in {{Neural Information Processing Systems}}},
  volume~36, pp.\  27147--27166, December 2023.

\bibitem[Zhao et~al.(2023)Zhao, Zhao, Boney, Kannala, and
  Pajarinen]{zhaoSimplifiedTemporalConsistency2023}
Yi~Zhao, Wenshuai Zhao, Rinu Boney, Juho Kannala, and Joni Pajarinen.
\newblock Simplified {{Temporal Consistency Reinforcement Learning}}.
\newblock In \emph{Proceedings of the 40th {{International Conference}} on
  {{Machine Learning}}}, pp.\  42227--42246. PMLR, July 2023.

\bibitem[Zhao et~al.(2025)Zhao, Scannell, Hou, Cui, Chen, Solin, Kannala, and
  Pajarinen]{ZhaoLeveraging2024}
Yi~Zhao, Aidan Scannell, Yuxin Hou, Tianyu Cui, Le~Chen, Arno Solin, Juho
  Kannala, and Joni Pajarinen.
\newblock Generalist world model pre-training for efficient reinforcement
  learning.
\newblock \emph{arxiv preprint arXiv:2502.19544}, 2025.

\end{thebibliography}
